\documentclass[11pt]{article}

\usepackage[final]{acl}

\usepackage{times}
\usepackage{latexsym}

\usepackage[T1]{fontenc}
\usepackage[utf8]{inputenc}
\usepackage{microtype}
\usepackage{inconsolata}
\usepackage{graphicx}
\usepackage{booktabs}
\usepackage{multirow}
\usepackage{tabularx}
\usepackage{makecell}
\usepackage{threeparttable}
\usepackage{adjustbox}
\usepackage{amsmath}
\usepackage[table]{xcolor}
\usepackage{amssymb}
\usepackage{pifont}
\usepackage[most]{tcolorbox}
\usepackage{listings}
\usepackage[table]{xcolor}
\definecolor{lightblue}{RGB}{235,242,250}
\newcommand{\gx}{\textcolor{red!50!black}{\xmark}}
\newcommand{\gc}{\textcolor{green!50!black}{\cmark}}
\usepackage{enumitem}
\usepackage{float}
\newcommand{\cmark}{\ding{51}}
\newcommand{\xmark}{\ding{55}}
\usepackage{subcaption}
\usepackage{kotex}
\title{CompCQR: Compositional Query Generation for Training-Free Conversational Search}

\author{Yunah Jang$^{1}$, Kang-il Lee$^{1}$, Joongbo Shin$^{2}$ and Kyomin Jung$^{1, \dagger}$\\
  $^{1}$Dept. of ECE, Seoul National University,
  $^{2}$LG AI Research
  \\
  \texttt{\{vn2209, 4bkang, kjung\}@snu.ac.kr}\\
  \texttt{jb.shin@lgresearch.ai}
  }
\begin{document}
\maketitle

\begingroup
\renewcommand{\thefootnote}{\fnsymbol{footnote}}
\setcounter{footnote}{1}
\footnotetext[2]{Corresponding author.}
\footnotetext[1]{Code is available at \url{https://github.com/YunahJang/CompCQR}.}
\endgroup

\begin{abstract}
Multi-turn interactions with LLMs are becoming increasingly common in information-seeking scenarios.
However, user queries are often ambiguous and context-dependent, making them ill-suited for direct use as retriever queries.
Conversational query reformulation (CQR) addresses this issue by rewriting the current utterance into a stand-alone query grounded in the dialogue history. 
Recent LLM-based CQR approaches achieve strong performance; however, their repeated LLM invocations and misalignment with downstream retrievers remain challenges.
In this work, we begin from the observation that retrievers are highly sensitive to content ordering: simply reordering the same content can lead to changes in retrieval coverage and performance.
Based on this, we propose a novel training-free method that generates a very large number of queries with minimal LLM usage by compositionally combining a small set of atomic components.
We further apply LLM reasoning to construct a high-quality document set that balances precision and recall while capturing the user’s core intent.
Our framework generalizes across both open- and closed-source LLMs as well as dense and sparse retrievers. 
It achieves strong performance on four widely used conversational benchmarks, with up to 22.5\% relative MRR improvement over the previous state-of-the-art baseline with far fewer LLM calls.$^{*}$

\end{abstract}

\section{Introduction}
Large language models (LLMs) are increasingly becoming a common medium for information seeking~\cite{searcho1, websearch}.
When interacting with LLMs, users often engage in multi-turn exchanges to refine and clarify their information needs~\cite{llm_lost, multi-turn, multi-turn_llm}.
This makes conversational search—the task of retrieving relevant information throughout an ongoing dialogue while understanding user intent in context—a fundamental paradigm for modern information access~\cite{survey, agentic_conversational_search}.

\begin{figure}[t]
    \centering
    \includegraphics[width=\linewidth]{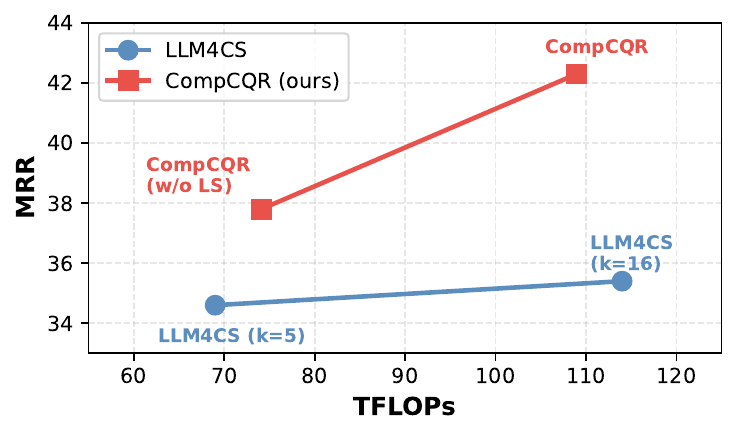}
    \caption{\textbf{Comparison of MRR across different FLOPs.} On TopiOCQA~\cite{topiocqa}, we conduct a comparison of two prompting-based CQR methods: CompCQR (ours) and LLM4CS~\cite{llm4cs}.}
    \label{fig:cost}
\end{figure}

In conversational settings, user utterances are often short, ambiguous, and highly dependent on prior dialogue context~\cite{topiocqa, qrecc, cast19}. They frequently contain ellipsis, coreference, and underspecified intent, making them difficult to use directly for retrieval~\cite{itercqr, convgqr}. To address these challenges, conversational query rewriting (CQR) reformulates the current utterance into a stand-alone query grounded in the dialogue history, enabling downstream retrievers to better capture the user’s information need~\cite{cqr, conqrr}.

LLMs have recently emerged as a powerful tool for CQR, owing to their strong performance to infer missing context from dialogue history and produce coherent stand-alone queries~\cite{adarewriter, llm4cs, llm-aided}. 
Despite their strong ability, important challenges remain in making methods both computationally efficient and effective.
In practice, as shown in Table~\ref{tab:method_comparison}, existing methods depend on multiple LLM calls or training models.

\begin{table}[t]
\centering
\small
\setlength{\tabcolsep}{4.8pt}
\begin{threeparttable}
\begin{adjustbox}{width=\linewidth}
\begin{tabular}{lcccc}
\toprule[1.5pt]
\shortstack{Method} 
& \shortstack{\# of\\Calls} 
& \shortstack{\# of\\Query} 
& \shortstack{Query \\Agg.}
& \shortstack{Train\\Free} \\
\midrule
T5QR~\citeyearpar{t5qr} & 0 & 1 & \gx & \gx \\
ConvGQR~\citeyearpar{convgqr} & 0 & 1 & \gx & \gx \\
IterCQR~\citeyearpar{itercqr} & 0 & 1 & \gx & \gx \\
ConvSearch-R1~\citeyearpar{convsearchr1} & 1 & 1 & \gx & \gx \\
AdaCQR~\citeyearpar{adacqr} & 1 & 1 & \gx & \gx \\
CHIQ-Fusion~\citeyearpar{chiq} & 6 & 2 & D & \gx \\
AdaRewriter~\citeyearpar{adarewriter} & 16 & 16 & \gx & \gx \\
LLM4CS~\citeyearpar{llm4cs} & 5 & 5 & Q & \gc \\
\midrule
\rowcolor{cyan!10}
CompCQR (Ours) & 3 & \textbf{325} & D & \gc \\
\bottomrule[1.5pt]
\end{tabular}
\end{adjustbox}
\caption{
\textbf{Comparison of CQR methods.}
We compare the number of LLM calls, the number of query reformulations, the aggregation (agg.) strategy, and whether the method is training-free. Aggregation occurs either at the query level (Q), merging multiple queries into a single query embedding, or at the document level (D), retrieving with each query and aggregating the document sets.
}
\label{tab:method_comparison}
\end{threeparttable}
\end{table}
Instead of relying on expensive training or excessive LLM inference, we propose CompCQR, a training-free CQR framework.
CompCQR generates diverse queries through lightweight compositional expansion with minimal LLM usage, enabling broader exploration of the retrieval space. 
It then aggregates the documents retrieved by these queries into a ranked list, improving coverage by retaining documents that may be missed by any single rewrite while preserving strong ranking signals from highly ranked results.
Finally, it applies LLM-based reasoning to select a document based on answerability, which constructs a document set that better balances recall and precision while remaining faithful to the user's intent.
As shown in Figure~\ref{fig:cost}, CompCQR yields substantially better retrieval performance than prompting-based baselines under a comparable computational cost.

We conduct extensive experiments on four conversational search benchmarks and find that CompCQR achieves competitive or superior performance over training-based QR baselines with a more LLM-efficient pipeline. In particular, it outperforms the previous best-performing baseline by up to 22.5\% on dense retrievers and 14.7\% on sparse retrievers. These results highlight the strong performance and robustness of the proposed method across diverse retrieval settings.

In summary, our contributions are as follows:
\vspace{-1mm}
\begin{itemize}[itemsep=2pt]
    \item We propose CompCQR, a training-free method that generates a large number of compositional rewrites with minimal LLM calls.
    \item We use document-level aggregation and an LLM-based reasoning step that provides both recall and precision improvements to the retrieved document set. 
    \item We demonstrate that CompCQR consistently performs well across multiple datasets, retrievers, and long conversational contexts.
\end{itemize}

\section{Potential of Compositional Queries}
\label{sec:preliminary}
Several works explore test-time scaling for CQR by generating multiple rewritten queries, resulting in an expanded retrieval space and improved retrieval performance~\cite{llm4cs, adarewriter}. 
However, such approaches typically require multiple LLM calls, resulting in substantial computational cost.
More broadly, recent work on query robustness has shown that even semantically equivalent query formulations can lead to substantially different retrieval behavior~\cite{howyouaskmatters}, suggesting that diverse formulations can expose complementary retrieval results.

Motivated by these observations, we propose a lightweight query generation method, compositional query generation. This method constructs multiple query variants solely by adding, removing, or reordering atomic components. 
Specifically, we extract four components using two LLM calls: a rewritten query–answer pair ($q^{r}, a^{r}$) and a follow-up query–answer pair ($q^{f}, a^{f}$). Together with the original query ($q$), this yields five components in total.
We then enumerate all permutations of these components to construct 325 candidate queries.\footnote{The total number of permutations from five components is $\sum_{r=1}^{5} {}_{5}P_{r}=325$.}

In this section, we show that compositional queries (i) yield diverse retrieval spaces, and  (ii) perform well compared to existing multiple-query rewriting methods. 
We employ the ANCE~\cite{ance} retriever and TopiOCQA~\cite{topiocqa} dataset.

\subsection{Retrieval Coverage of Compositional Queries}
% \subsection{Retriever Sensitivity to Component Order}
\label{sec:retrieval_coverage}
We first investigate whether compositional queries yield diverse retrieval results. 
Since compositional queries are formed by reordering the same set of components, many variants differ only in their ordering. 
We therefore compare queries with identical components but different orders, and measure the overlap of their top-k retrieved documents using Jaccard similarity. 
The average overlap is 0.72 at top-10 and 0.71 at top-100. These results indicate that, even with identical components, changing their order can result in a Jaccard distance of 28\%, reflecting a substantial set-level difference in the retrieved documents. This suggests that compositional queries can improve retrieval coverage by exploiting the retriever's sensitivity to query order.

This trend is consistent across different dense retrievers, as shown in Appendix~\ref{sec:different_retrievers}.

\begin{figure}[t]
    \centering
    \includegraphics[width=0.9\columnwidth]{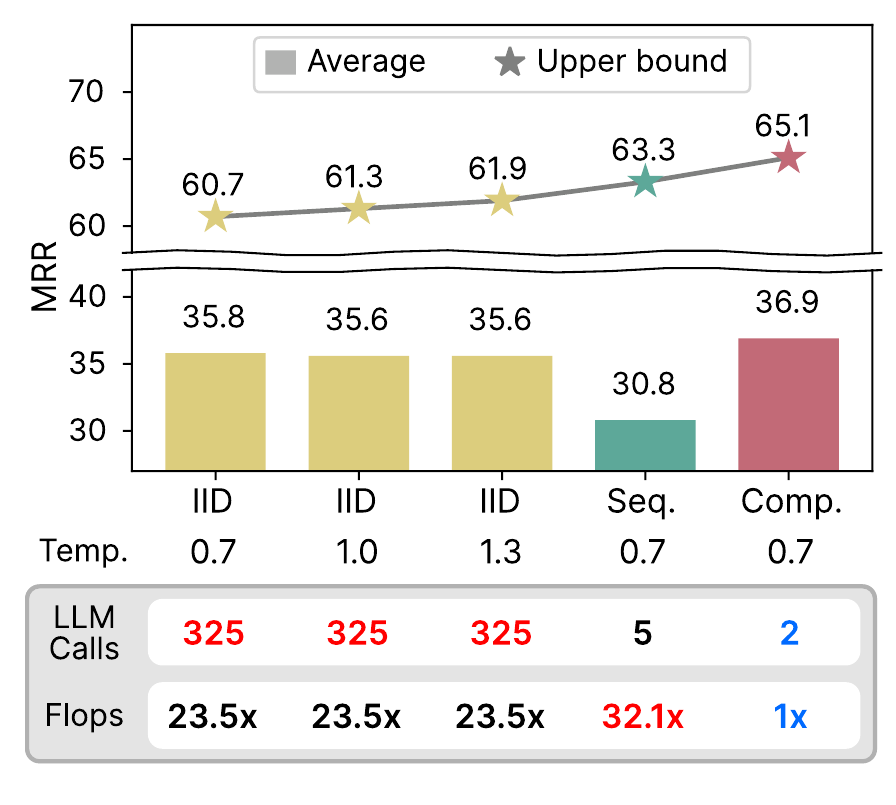}
\caption{\textbf{Comparison of query rewriting methods.} 
We compare the average and upper-bound retrieval performance across 325 query candidates generated by compositional and sampling-based query rewriting methods, where IID denotes independent sampling at different temperatures, Seq.\ sequential sampling, and Comp.\ compositional query rewriting.}
\vspace{-2mm}
\label{fig:preliminaryu}
\end{figure}

\subsection{Effectiveness of Compositional Queries}
We then investigate whether the retrieval diversity of compositional queries leads to higher CQR performance compared to existing methods. 
For existing multiple-query generation approaches, we adopt two LLM sampling strategies: IID sampling at three temperatures (0.7, 1.0, and 1.3), and sequential sampling with five calls, each generating 65 queries conditioned on previously generated outputs. 
For a fair comparison, all methods generate 325 queries per instance over 100 randomly sampled instances from TopiOCQA~\cite{topiocqa}.

As shown in Figure~\ref{fig:preliminaryu}, compositional query rewriting achieves the highest average and upper-bound MRR while requiring only two LLM calls per query, compared to 325 calls for IID sampling and 32.1$\times$ higher FLOPs for sequential generation. This result underscores a key insight: the goal of rewriting queries is not merely to produce uniformly strong candidates, but to discover reformulations that better align with the retriever and recover relevant documents missed by the original query.

\section{Methodology}
\begin{figure*}[t]
    \centering
    \includegraphics[width=0.9\linewidth]{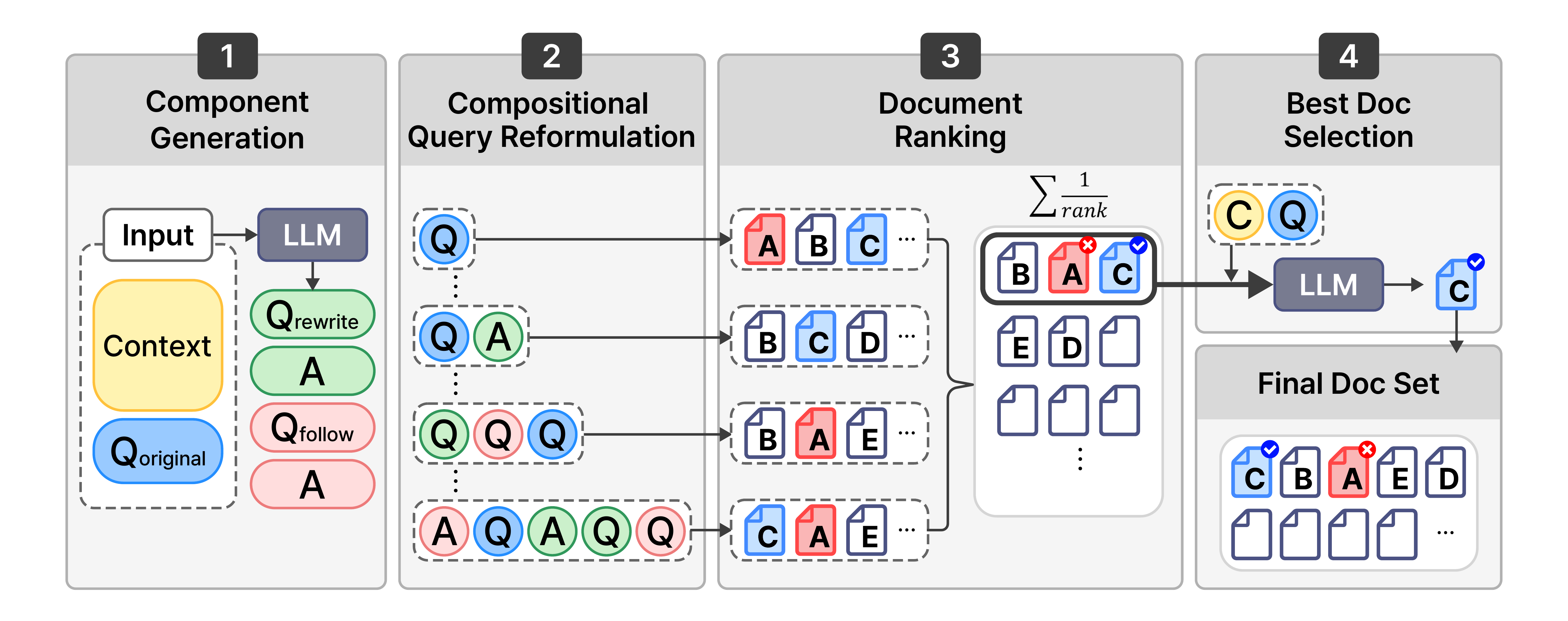}
    \caption{\textbf{CompCQR Framework Overview.} CompCQR first extracts query components with two LLM calls, generates compositional query candidates, aggregates retrieved documents via RRF scoring, and finally performs LLM-based document selection.}
    \label{fig:overview}
\end{figure*}

In this section, as illustrated in Figure~\ref{fig:overview}, we introduce a three-stage framework designed to improve document relevance through complementary recall- and precision-oriented components. Specifically, the recall-oriented stages consist of compositional query generation and document ranking, while the precision-oriented stage is implemented through LLM-based document selection.

\subsection{Task Formulation}
A conversational search task aims to retrieve documents that contain the necessary information to answer a user's query in a conversational setting.
At the current turn $t$, the user issues a query $q_t$, and the historical context $H_{t-1}$ is defined as 
\[
H_{t-1} = \{(q_i, a_i)\}_{i=1}^{t-1},
\]
where $q_i$ and $a_i$ denote the query and answer at turn $i$, respectively.
Given the current query $q_t$ and the historical context $H_{t-1}$, a language model $\mathcal{M}$ rewrites $q_t$ into a standalone query:
\[
\tilde{q_t} = \mathcal{M}(q_t, H_{t-1}).
\]

The rewritten query $\tilde{q_t}$ is then fed into an off-the-shelf retriever $\mathcal{R}$, which returns a ranked list of documents:
\[
D_t = \mathcal{R}(\tilde{q_t}) = \{d_1, d_2, \dots, d_k\},
\]
where $D_t$ denotes the top-$k$ retrieved documents.  
In later sections, we denote the history and current query as $H$ and $q$, respectively.

\subsection{Compositional Query Reformulation}
\label{sec:compositonal_query_expansion}

As discussed in Section \ref{sec:preliminary}, we employ compositional query reformulation to broaden retrieval coverage. For each instance, we first use dialogue history $H$ and original query $q$ to issue two LLM calls with different prompts, and then construct multiple compositional query variants from their outputs. Unlike prior work, which primarily considers rewrites and potential answers, we additionally generate a follow-up query and its corresponding answer, further increasing semantic diversity~\cite{proactive_benchmark, followup_query}.

Specifically, using the query rewriting prompt $P_{\mathrm{rewrite}}$ and the follow-up query generation prompt $P_{\mathrm{follow}}$, we obtain four components: a rewritten query $\tilde{q}$ and its potential answer $\tilde{a}$, and a follow-up query $q^{\mathrm{f}}$ and its corresponding answer $a^{\mathrm{f}}$:
\begin{align}
(\tilde{q}, \tilde{a}) &= LLM(P_{\mathrm{rewrite}}, H, q), \\
(q^{\mathrm{f}}, a^{\mathrm{f}}) &= LLM(P_{\mathrm{follow}}, H, q).
\end{align}

Together with the original query $q$, these outputs form five query-related components:
\begin{equation}
\mathcal{C} = \{ q,\ \tilde{q},\ \tilde{a},\ q^{\mathrm{f}},\ a^{\mathrm{f}} \}.
\end{equation}
We then construct compositional query candidates by enumerating all possible ordered combinations of these five components. Specifically, for each length $r \in \{1, 2, 3, 4, 5\}$, we consider all permutations without repetition:
\begin{equation}
\mathcal{E}
=
\bigcup_{r=1}^{5} \mathrm{Perm}(\mathcal{C}, r), \quad |\mathcal{E}| = \sum_{r=1}^{5} {}_{5}P_{r}
= 325.
\end{equation}
where $\mathrm{Perm}(\mathcal{C}, r)$ denotes the set of all length-$r$ ordered sequences formed from $\mathcal{C}$.\footnote{For BM25 retrieval, component order does not affect retrieval because queries are treated as bags of words. Therefore, the number of distinct query combinations is $\sum_{r=1}^{5} {5 \choose r} = 31$.} The prompts used in this step are provided in Appendix~\ref{sec:prompt}.

\subsection{Document Sorting by Retrieval Scores}

Using multiple queries to explore the query space has shown promise in prior work~\cite{adarewriter, itercqr}. Existing approaches typically combine query information at the query or embedding level~\cite{llm4cs, cmqr}, while Reciprocal Rank Fusion (RRF) is often used as a learning signal in retrieval pipelines~\cite{adarewriter, adacqr}. In contrast, we apply RRF directly to document-set-level aggregation~\cite{chiq, cdr_lin, genqrensemble}. By aggregating documents retrieved from multiple queries using RRF~\cite{rrf} together with document frequency, our method preserves complementary evidence across queries and thereby improves document coverage.

\begin{table*}[t]
\centering
\small
\begin{threeparttable}

\begin{adjustbox}{max width=\textwidth}
\begin{tabular}{llcccccccccc}
\toprule[1.2pt]
\multirow{2}{*}{\makecell{\rule{0pt}{2.2ex}}} & \multirow{2}{*}{\makecell{\rule{0pt}{3ex}\textbf{Method}}} & \multirow{2}{*}{\makecell{\rule{0pt}{3ex}\textbf{Model}}} & \multirow{2}{*}{\makecell{\rule{0pt}{3ex}\textbf{TF}}}
& \multicolumn{4}{c}{\textbf{QReCC}}
& \multicolumn{4}{c}{\textbf{TopiOCQA}} \\
\cmidrule(lr){5-8} \cmidrule(lr){9-12}
& & & & \textbf{MRR} & \textbf{NDCG} & \textbf{R@10} & \textbf{R@100}
& \textbf{MRR} & \textbf{NDCG} & \textbf{R@10} & \textbf{R@100} \\
\midrule

\multirow{16}{*}{\rotatebox[origin=c]{90}{\textbf{{Sparse (BM25)}}}}
& T5QR & T5-base & \gx & 33.4 & 30.2 & 53.8 & 86.1 & 11.3 & 9.8 & 22.1 & 44.7 \\
& \textsc{ConQRR} & T5-base & \gx & 38.3 & - & 60.1 & 88.9 & - & - & - & - \\
& ConvGQR & T5-base & \gx & 44.1 & 41.0 & 64.4 & 88.0 & 12.4 & 10.7 & 23.8 & 45.6 \\
& \textsc{EdiRCS} & T5-base & \gx & 42.1 & - & 65.6 & 85.3 & - & - & - & - \\
& IterCQR & T5-base & \gx & 46.7 & 44.1 & 64.4 & 85.5 & 16.5 & 14.9 & 29.3 & 54.1 \\
& ReTPO & Llama-2-7B & \gx & 50.0 & 47.3 & 69.5 & 89.5 & 28.3 & 26.5 & 48.3 & 73.1 \\
& ICR & Llama2-7B & \gx & \underline{58.4} & \underline{54.9} & 78.3 & \underline{95.9} & 31.4 & 30.4 & 52.8 & 76.3 \\
& ConvSearch-R1 & Llama-3.2-3B & \gx & 55.9 & 54.3 & 77.2 & 89.0 & \underline{37.8} & \underline{36.2} & \underline{59.6} & \underline{80.1} \\
& ADACQR &  Llama-2-7B$^\dagger$ & \gx & 55.1 & 52.5 & 76.5 & - & 28.3 & 26.5 & 48.9 & - \\
& CHIQ-Fusion &  Llama-2-7B$^\dagger$ & \gx & 54.3 & 51.9 & 78.5 & - & 25.6 & 23.5 & 44.7 & - \\
& AdaRewriter & Llama-3.1-8B$^\dagger$  & \gx & 56.2 & 53.8 & 78.8 & - & 30.7 & 28.8 & 51.3 & - \\
\cmidrule{2-12}
& LLM-Aided & GPT-3.5-turbo & \gc & 49.4 & 46.5 & 67.1 & - & - & - & - & - \\
& LLM4CS & Llama-3.1-8B & \gc & 49.7 & 46.9 & 73.8 & - & 24.5 & 22.6 & 42.1 & - \\
& LLM4CS& GPT-4.1-mini & \gc & 54.4 & 52.0 & \underline{78.9} & 94.7 & 31.3 & 29.6 & 52.7 & \underline{74.9} \\
\rowcolor{cyan!10}
& CompCQR & Llama-3.1-8B & \gc & 55.7 & 54.0 & 75.8 & 94.1 & 29.3 & 28.4 & 49.3 & 73.7 \\
\rowcolor{cyan!10}
& CompCQR & GPT-4.1-mini & \gc & \textbf{67.0} & \textbf{64.4} & \textbf{83.4} & \textbf{96.1} & \textbf{41.3} & \textbf{40.4} & \textbf{60.3} & \textbf{80.2} \\

\midrule

\multirow{13}{*}{\rotatebox[origin=c]{90}{\textbf{{Dense (ANCE)}}}}
& T5QR & T5-base & \gx & 34.5 & 31.8 & 53.1 & 72.8 & 23.0 & 22.2 & 37.6 & 54.4 \\
& ConvGQR & T5-base & \gx & 42.0 & 39.1 & 63.5 & 81.8 & 25.6 & 24.3 & 41.8 & 58.8 \\
& IterCQR & T5-base & \gx & 42.9 & 40.2 & 65.5 & 84.1 & 26.3 & 25.1 & 42.6 & 62.0 \\
& ReTPO & Llama-2-7B & \gx & 44.0 & 41.1 & 66.7 & 84.6 & 30.0 & 28.9 & 49.6 & 68.7 \\
& ICR & Llama2-7B & \gx & 49.5 & 46.8 & 73.2 & \textbf{88.3} & 42.1 & 40.4 & 64.3 & 78.3 \\
& ConvSearch-R1 & Llama-3.2-3B & \gx & 50.2 & 48.1 & \underline{70.6} & 82.8 & \textbf{50.5} & \textbf{50.1} & \textbf{72.0} & \textbf{86.3} \\
& ADACQR & Llama-2-7B$^\dagger$  & \gx & 45.8 & 42.9 & 67.3 & - & 38.5 & 37.6 & 58.4 & - \\
& CHIQ-Fusion & Llama-2-7B$^\dagger$  & \gx & 47.2 & 44.2 & 70.7 & - & 38.0 & 37.0 & 61.6 & - \\
& AdaRewriter & Llama-3.1-8B$^\dagger$  & \gx & 47.5 & 44.7 & 69.8 & - & 40.3 & 39.7 & 61.9 & - \\
\cmidrule{2-12}
& LLM-Aided & GPT-3.5-turbo & \gc & 43.5 & 41.3 & 65.6 & - & - & - & - & - \\
& LLM4CS & Llama-3.1-8B & \gc & 43.2 & 40.7 & 64.6 & - & 35.2 & 34.8 & 58.2 & 76.0 \\
& LLM4CS & GPT-4.1-mini & \gc & 47.4 & 44.6 & 70.6 & 87.4 & 41.9 & 40.9 & 63.7 & 78.7 \\
\rowcolor{cyan!10}
& CompCQR & Llama-3.1-8B & \gc & \underline{50.8} & \underline{48.3} &  67.8 & 86.2 & 42.3 & 41.5 & 62.2 & 80.2 \\
\rowcolor{cyan!10}
& CompCQR & GPT-4.1-mini & \gc & \textbf{61.5} & \textbf{58.6} & \textbf{74.6} & \underline{88.0}& \underline{48.7} & \underline{47.7} & \underline{67.5} & \underline{83.5} \\

\bottomrule[1.2pt]
\end{tabular}
\end{adjustbox}
\caption{
\textbf{Main results on QReCC and TopiOCQA.} We report our main results under sparse (BM25) and dense (ANCE) retrieval.
We compare our proposed method, {\setlength{\fboxsep}{0.5pt}\colorbox{cyan!12}{\strut CompCQR}}, with prior conversational query rewriting approaches.
\textbf{Bold} and \underline{underlined} values denote the best and second-best results within each retriever setting.
$^\dagger$ denotes methods using fine-tuned LMs with LLM. TF stands for training-free.}
\label{table:main_results}

\end{threeparttable}
\end{table*}

For a compositional query set $\mathcal{E}$, let $\mathrm{rank}_e(d)$ denote the rank of document $d$ among the top 100 retrieved documents for query candidate $e \in \mathcal{E}$. We compute the RRF score of document $d$ as
\begin{equation}
\mathrm{RRF}(d) = \sum_{e \in \mathcal{E}_d} \frac{1}{s + \mathrm{rank}_e(d)},
\end{equation}
where $\mathcal{E}_d \subseteq \mathcal{E}$ denotes the subset of query candidates that retrieve $d$, and $s$ is a smoothing hyperparameter. Documents are then ranked in descending order of $\mathrm{RRF}(d)$. Ties are broken using $\mathrm{Freq}(d)=|\mathcal{E}_d|$, giving higher priority to documents retrieved by more query candidates.

This aggregation favors documents that appear frequently and at high ranks across query candidates, capturing the user's information need. By exposing the retriever to multiple views of the same need, it increases the likelihood of retrieving relevant documents missed by the original query.

\subsection{Best Document Selection}
Finally, we refine the top-ranked result by leveraging the LLM's document-grounded reasoning capabilities. 
Let $\hat{D}_{100} = (d_1, d_2, \dots, d_{100})$ denote the top-100 documents ranked by RRF-based aggregation, and let $\hat{D}_{k} = (d_1, \dots, d_k)$ denote its top-$k$ prefix. 
The LLM then selects a single document
\begin{equation}
d^{*} = \mathcal{M}(q, H, \hat{D}_{k}),
\end{equation}
where $d^{*} \in \hat{D}_{k}$ is judged most likely to contain answer-supporting information. The final list $D^{\mathrm{final}}_{100}$ is constructed by moving $d^{*}$ to rank 1 while preserving the relative order of the remaining documents in $\hat{D}_{100}$. This list is used for evaluation.

\section{Experiments}
\subsection{Experimental Setup}
\paragraph{Datasets.} We evaluate our framework on four widely used conversational search datasets: QReCC~\cite{qrecc}, TopiOCQA~\cite{topiocqa}, CAsT19~\cite{cast19}, and CAsT20~\cite{cast20}. 
We evaluate baselines on the test set, considering only instances that contain gold annotations. 
Additional statistics are provided in Appendix~\ref{sec:statistics}.

\paragraph{Evaluation Metrics.}
We use the pytrec\_eval toolkit to compute retrieval metrics~\cite{pytrec_eval}, including mean reciprocal rank (MRR), NDCG@3, Recall@10, and Recall@100, following previous studies~\cite{adarewriter, itercqr, convgqr}.

\paragraph{Implementation Details.}
We evaluate our framework using two types of models: Llama-3.1-8B-Instruct~\cite{grattafiori2024llama} and GPT-4.1-mini~\cite{gpt4.1}. 
Llama-3.1-8B has been widely used in prior baselines, often together with additional reward model training. 
ChatGPT has also been commonly adopted in prompting-based approaches; therefore, we use GPT-4.1-mini as a lightweight yet strong-performing model for comparison with these baselines. 
For RRF, we tune the hyperparameters on the validation set, setting $s$ to 0 and choosing top-$K$ values of 10 for Llama-3.1-8B and 40 for GPT-4.1-mini. 
Additional analyses of these hyperparameters are provided in Appendix~\ref{app:hyperparameters}. 
For retrieval, we use both BM25~\cite{bm25} and the ANCE~\cite{ance} dense retriever, following previous work~\cite{adacqr, adarewriter, retpo, itercqr}.

\paragraph{Baselines.}
We compare our method against three types of baselines: training-based methods, hybrid methods that combine LLMs with trained smaller models, and training-free LLM-based methods. 
The training-based baselines include T5QR~\cite{t5qr}, CONQRR~\cite{conqrr}, ConvGQR~\cite{convgqr}, EdiRCS~\cite{edirics}, IterCQR~\cite{itercqr}, RetPO~\cite{retpo}, ICR~\cite{icr}, and ConvSearch-R1~\cite{convsearchr1}. 
Among hybrid methods, which combine trained language models such as T5-base or DeBERTa with LLMs, we include AdaCQR~\cite{adacqr}, CHIQ-Fusion~\cite{chiq}, and AdaRewriter~\cite{adarewriter}. For the training-free setting most comparable to ours, we consider LLM-Aided~\cite{llm-aided} and LLM4CS~\cite{llm4cs}. Further baseline details are provided in Appendix~\ref{sec:baseline_details}.

\begin{table}[t]
\centering
\small
\begin{threeparttable}

\begin{adjustbox}{width=\columnwidth}
\begin{tabular}{lcccc}
\toprule[1.2pt]
\multirow{2}{*}{\makecell{\rule{0pt}{3ex}\textbf{Method}}}
& \multicolumn{2}{c}{\textbf{CAsT-19}}
& \multicolumn{2}{c}{\textbf{CAsT-20}} \\
\cmidrule(lr){2-3} \cmidrule(lr){4-5}
& \textbf{MRR} & \textbf{R@10}
& \textbf{MRR} & \textbf{R@10}  \\
\midrule

LLM4CS$^{*}$         
& \underline{73.3} 
& \textbf{13.9} 
& \underline{77.2} 
& \underline{17.8} 
\\

AdaRewriter$^{\circ}$     
& - 
& 13.0 
& 63.0 
& \textbf{21.6} 
\\

\rowcolor{cyan!10}
CompCQR$^{\circ}$ 
& 66.6 
& 12.1 
& 71.0 
& 16.2 
\\

\rowcolor{cyan!10}
CompCQR$^{*}$ 
& \textbf{75.8} 
& \underline{13.2} 
& \textbf{78.3} 
& 16.7 
\\

\bottomrule[1.2pt]
\end{tabular}
\end{adjustbox}

\caption{
\textbf{CAsT-19 and -20 results.}
We compare {\setlength{\fboxsep}{0.5pt}\colorbox{cyan!12}{\strut CompCQR}} against prior baselines under dense retrieval.
$^{\circ}$ denotes results using Llama-3.1-8B-Instruct, and
$^{*}$ denotes results using GPT-4.1-mini.
}
\label{tab:cast}

\end{threeparttable}
\end{table}

\subsection{Main Results}
In Table~\ref{table:main_results}, we evaluate TopiOCQA and QReCC using both dense (ANCE) and sparse (BM25) retrievers. Compared with training-based baselines that require human-rewritten queries or ground-truth documents, our method achieves comparable performance and, in some settings, outperforms them by up to 22.5\% (+11.3 MRR) without using any training labels. It also outperforms all prior training-free LLM-based query rewriting methods by a large margin across all settings. We also evaluate on CAsT-19 and CAsT-20 (Table~\ref{tab:cast}), where our approach outperforms existing prompting-based methods. Overall, these results show that CompCQR achieves strong performance without explicit training, outperforming previous baselines while reducing LLM inferences (see Appendix~\ref{app:computational_cost}).

\section{Analysis}

\subsection{Ablation Studies}
\begin{table}[t]
\centering
\small
\begin{threeparttable}

\begin{adjustbox}{width=\linewidth}
\begin{tabular}{lccc}
\toprule[1.2pt]
\textbf{Setting} & \textbf{MRR} & \textbf{NDCG} & \textbf{R@10} \\
\midrule

CompCQR (Ours)               & \textbf{42.3} & \textbf{41.5} & \textbf{62.2} \\
\quad - w/o RRF (+Rel.)    & 35.7       &  34.4       &  51.5        \\
\quad - w/o RRF (+Mean)    & 39.4          & 38.9         & 57.9          \\
\quad - w/o LS          & 37.8          & 36.5          & \textbf{62.2} \\
\quad - w/o RRF (+Rel.) \& LS   & 31.5        & 29.9          & 51.5          \\
\quad - w/o RRF (+Mean) \& LS  & 36.8       & 35.8      &    57.9       \\
\bottomrule[1.2pt]
\end{tabular}
\end{adjustbox}
\caption{
\textbf{Ablation study.}
We evaluate the contribution of the two core components of CompCQR: LLM-based document selection (LS) and RRF-based document aggregation (RRF).
For the setting w/o RRF, candidate passages are ranked either according to the retriever relevance score (Rel.) or using mean aggregation (Mean).
}
\label{tab:ablation}
\end{threeparttable}
\end{table}

\begin{table}[t]
\centering
\small
\resizebox{\columnwidth}{!}{%
\begin{tabular}{cccccc}
\toprule[1.2pt]
\multirow{2}{*}{\textbf{\#Q}} & \multirow{2}{*}{\textbf{LS}} 
& \multicolumn{2}{c}{\textbf{RRF}} 
& \multicolumn{2}{c}{\textbf{Mean}} \\
\cmidrule(lr){3-4} \cmidrule(lr){5-6}
& & \textbf{MRR} & \textbf{NDCG} & \textbf{MRR} & \textbf{NDCG} \\
\midrule
\multirow{2}{*}{5} 
  & \ding{55} & 36.0 & 34.6 & 34.2 & 33.1 \\
  & \ding{51} & 39.6{\scriptsize\,(+3.6)} & 38.5{\scriptsize\,(+3.9)} & 38.1{\scriptsize\,(+3.9)} & 37.5{\scriptsize\,(+4.4)} \\
\midrule
\multirow{2}{*}{325}
  & \ding{55} & 37.8 & 36.5 & 36.8 & 35.8 \\
  & \ding{51} & \textbf{42.3}{\scriptsize\,(+4.5)} & \textbf{41.5}{\scriptsize\,(+5.0)} & 39.4{\scriptsize\,(+2.6)} & 38.9{\scriptsize\,(+3.1)} \\
\bottomrule[1.2pt]
\end{tabular}%
}
\caption{\textbf{Comparison with different aggregation strategies.} We compare mean and RRF aggregation with and without the LLM selection (LS) setting on the TopiOCQA dataset. Each method is evaluated using 5 and 325 queries. The values in parentheses indicate the performance gain when LS is applied.}
\label{tab:aggregation_strategy_compare}
\end{table}

We conduct an ablation study on TopiOCQA using Llama-3.1-8B-Instruct to assess the contribution of each component in CompCQR.
It consists of two main components: an RRF-based document aggregation stage (RRF) and a final LLM-based document selection stage (LS).

To examine the effect of removing the LLM-based selection module (w/o LS) we directly use the ranking produced by RRF aggregation.
To examine the contribution of the RRF-based aggregation module, we further construct \textit{w/o RRF}, which removes the RRF-based aggregation step.
In this setting, documents are instead aggregated using either the retriever relevance score (Rel.) or query embedding mean aggregation (Mean).
We note that query mean aggregation is also adopted in the strong baseline LLM4CS~\cite{llm4cs}.
The results in Table~\ref{tab:ablation} show that both components contribute substantially to retrieval quality, with a 4.5-point drop when LS is removed.

We further verify in Appendix~\ref{app:ls} that this gain does not simply come from the additional LLM call, as applying the same selection step to LLM4CS yields a much smaller improvement than for CompCQR.
Replacing top-1 selection with full reranking over the top-10 documents also degrades performance substantially (Appendix~\ref{app:llm_reranking}), indicating that selecting the best supporting document is a more reliable decision for the LLM than ordering an entire list.

\subsection{Aggregation Method under Scaling}
\label{sec:aggregation_method}

We further investigate the effect of aggregation strategies under different numbers of query candidates. 
Table~\ref{tab:aggregation_strategy_compare} compares reciprocal rank fusion (RRF) and mean aggregation using 5 randomly sampled query candidates and the full set of 325 query candidates, both with and without LLM selection.

Overall, RRF consistently outperforms mean aggregation across all settings. 
In particular, CompCQR with RRF improves by 3 points when scaling from 5 to 325 candidates in NDCG@3, whereas mean aggregation shows a smaller gain of 1.4 points. 
Moreover, at 325 candidates, incorporating LLM selection yields the largest gain of 5.0 NDCG@3 points over RRF aggregation alone.
These results suggest that document-level aggregation benefits more from LLM-based document selection than query-embedding averaging, especially when a large and diverse set of compositional queries is used.

\subsection{Computational Efficiency}
\begin{table}[t!]
\centering
\small
\begin{tabular}{lccc}
\toprule[1.2pt]
Method & \# Cands. & End-to-End (s) & MRR \\
\midrule
LLM4CS  & 5   & \textbf{4.05} & 34.6 \\
LLM4CS  & 16  & 5.51 & 35.4 \\
LLM4CS  & 325 & 29.90 & 40.5 \\
\midrule
CompCQR & 5   & 15.67 & 40.9 \\
CompCQR & 325 & 17.41 & \textbf{42.3} \\
\bottomrule[1.2pt]
\end{tabular}
\caption{End-to-end wall-clock latency and MRR on TopiOCQA across different numbers of candidates.}
\label{tab:latency}
\end{table}

\begin{table}[t!]
\centering
\small
\begin{tabular}{llrr}
\toprule [1.2pt]
Step & Description & Time & Fraction \\
\midrule
0 & Rewrite  & 2.93 s & 17.0\% \\
1 & Follow-up Generation & 1.80 s & 10.3\% \\
2 & Variation Generation & 0.19 ms & $\sim$0.0\% \\
3 & Retrieval + Sorting & 1.83 s & 10.5\% \\
4 & LLM-based Selection & 10.85 s & 62.3\% \\
\midrule
 & LLM Total & 15.58 s & 89.5\% \\
 & Retrieval Total &1.83 s & 10.5\% \\
\midrule
 & Overall & 17.41 s & 100.0\% \\
\bottomrule[1.2pt]
\end{tabular}
\caption{End-to-end latency breakdown of CompCQR on TopiOCQA.}
\label{tab:latency_breakdown}
\end{table}
We analyze the computational efficiency of CompCQR from two perspectives: end-to-end wall-clock latency, and the number of LLM calls and FLOPs per query.
Latency is measured on TopiOCQA using one NVIDIA A6000 GPU with vLLM.

Table~\ref{tab:latency} reports the average per-query latency together with MRR.
CompCQR consistently achieves higher retrieval effectiveness than LLM4CS across candidate sizes.
Notably, CompCQR with only 5 candidates achieves an MRR of 40.9, outperforming LLM4CS with 325 candidates (40.5) while being approximately 1.9$\times$ faster (15.67s vs.\ 29.90s).
With 325 candidates, our main setting, CompCQR further improves MRR to 42.3.

While CompCQR considers a large number of query candidates, the associated retrieval and sorting cost remains small.
As shown in Table~\ref{tab:latency_breakdown}, the end-to-end latency grows only marginally with the number of candidates, since retrieval and sorting account for just 10.5\% of the total runtime.
CompCQR is also favorable in terms of LLM calls and FLOPs, requiring only three calls per query (Appendix~\ref{app:computational_cost}).
These results show that CompCQR remains computationally efficient despite exploring a large candidate space.

\subsection{Analysis on Different Retrievers}
\begin{table}[t]
\centering
\small
\resizebox{\columnwidth}{!}{%
\begin{tabular}{llcc}
\toprule
\textbf{Method} & \textbf{Retriever} & \textbf{MRR} & \textbf{NDCG@3} \\
\midrule
LLM4CS  & GTR        & 42.6 & 41.7 \\
CompCQR & GTR        & 47.2 (+4.6) & 45.9 (+4.2) \\
\midrule
LLM4CS  & Contriever & 46.1 & 45.1 \\
CompCQR & Contriever & \textbf{48.9} (+2.8) & \textbf{47.6} (+2.5) \\
\bottomrule
\end{tabular}%
}
\caption{\textbf{Results with additional dense retrievers}. We report GTR and Contriever results on our framework.}
\label{tab:additional_dense_retrievers}
\end{table}

We conduct additional dense retriever experiments on TopiOCQA using two widely used dense retrievers, GTR~\cite{gtr} and Contriever~\cite{contriever}, with GPT-4.1-mini.

As shown in Table~\ref{tab:additional_dense_retrievers}, CompCQR consistently improves over LLM4CS in terms of MRR, with gains of +4.6 points using GTR and +2.8 points using Contriever. These results indicate that the effectiveness of CompCQR is not limited to ANCE, but generalizes to other dense retrievers.
We additionally analyze their order sensitivity in Appendix~\ref{sec:different_retrievers}.

\begin{figure}[t]
    \centering
    \includegraphics[width=0.85\columnwidth]{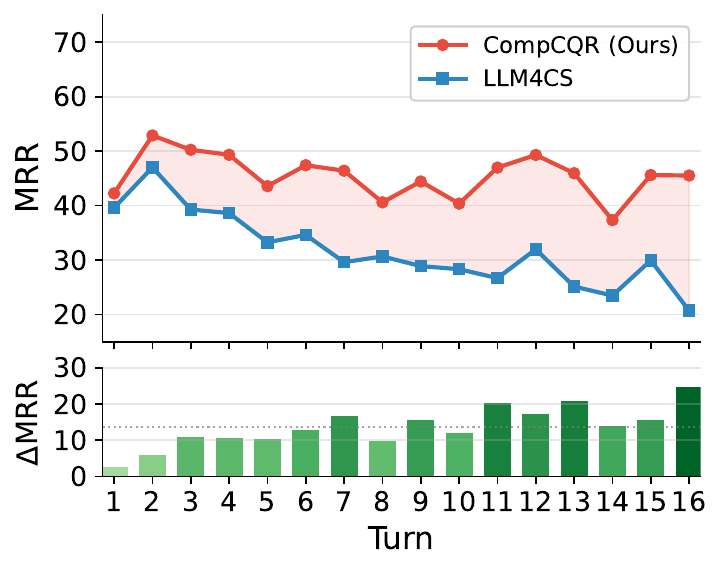}
    \caption{\textbf{Per-turn performance comparison.} We compare per-turn MRR with LLM4CS~\citeyearpar{llm4cs}.}
    \label{fig:turn_performance}
\end{figure}

\subsection{Long-conversation Performance}
Previous studies have shown that conversational search performance tends to decline in later turns~\cite{adarewriter, conqrr}. To evaluate the robustness of our framework in long conversations, we analyze performance by turn index. Figure~\ref{fig:turn_performance} compares our method with the LLM-prompting-based baseline LLM4CS~\cite{llm4cs} on TopiOCQA using GPT-4.1-mini. The performance gap widens as the conversation progresses, indicating that our framework is more robust in long conversations and better preserves retrieval effectiveness in later turns.

\section{Compositional Query Analysis}
In this section, we analyze our compositional query generation framework from three perspectives: (i) the relative contribution of each query component, (ii) differences in preferred supporting components across retrievers, and (iii) impact of component ordering on dense retrieval.

\begin{table}[t]
\resizebox{\columnwidth}{!}{%
\begin{tabular}{lccc}
\toprule[1.3pt]
\textbf{Component} & \textbf{Notation} & \textbf{Dense ($\phi$)} & \textbf{Sparse ($\phi$)} \\
\midrule
Original Query   & $q$              & 1.13  & 0.41 \\
Rewrite          & $\tilde{q}$      & \underline{9.98}  & 6.27 \\
Rewrite Answer   & $\tilde{a}$      & \textbf{16.20} & \textbf{14.54} \\
Followup Query   & $q^{\mathrm{f}}$ & 5.13  & 3.86 \\
Followup Answer  & $a^{\mathrm{f}}$ & 7.13  & \underline{6.32} \\
\bottomrule[1.3pt]
\end{tabular}%
}
\caption{\textbf{Impact of components by Shapley value.}
Shapley value analysis under dense and sparse retrievers, where all values are scaled by 100.
}
\label{tab:component_shapley_dense_sparse}
\end{table}
\begin{table}[t]
\centering
\begin{adjustbox}{width=\linewidth}
\begin{tabular}{lcccc}
\toprule[1.3pt]
 & \# Queries & MRR & NDCG@3 \\
\midrule
CompCQR (ours) & 325 & 42.3 & 41.5 \\
- without $q^f, a^f$ & 113 & 34.9 & 35.1\\
\bottomrule[1.3pt]
\end{tabular}
\end{adjustbox}
\caption{\textbf{Performance comparison without follow-up components.} 
We compare the full model with the variant without the follow-up query $q^f$ and answer $a^f$.}
\label{tab:follow_component}
\end{table}

\paragraph{Answer components are important.}
Our framework uses five components: the original query ($q$), rewritten query ($\tilde{q}$), rewritten answer ($\tilde{a}$), follow-up query ($q^{\mathrm{f}}$), and follow-up answer ($a^{\mathrm{f}}$).
To quantify the contribution of each component, we compute its \textit{Shapley value} on test-set MRR~\cite{shapely1,shapely2}, which measures the average marginal contribution of that component across all possible combinations.
As shown in Table ~\ref{tab:component_shapley_dense_sparse}, for both ANCE and BM25, the rewritten answer $\tilde{a}$ is the most important component, with Shapley values of 16.2 and 14.54, respectively.
This consistent pattern aligns with prior CQR studies showing that answer information is particularly useful for retrieval~\cite{convgqr,itercqr}.

In Table~\ref{tab:follow_component}, we additionally report the results after excluding  $q^{\mathrm{f}}$ and $a^{\mathrm{f}}$.
Removing these components leads to an overall MRR drop of 6.4 points.
This shows that follow-up information expands the query space with complementary intent signals, leading to more effective retrieval.

\paragraph{Different retrievers favor different components.}
The two retrievers differ, however, in their next most useful components.
In ANCE, the rewritten query $\tilde{q}$ is the second most influential component ($\phi = 9.98$), whereas in BM25, the follow-up answer $a^{\mathrm{f}}$ slightly surpasses $\tilde{q}$ ($\phi = 6.32$ vs.\ $\phi = 6.27$).
This suggests that sparse and dense retrievers benefit from different supporting components and composition strategies.

\begin{figure}[t]
    \centering
    \includegraphics[width=0.9\columnwidth]{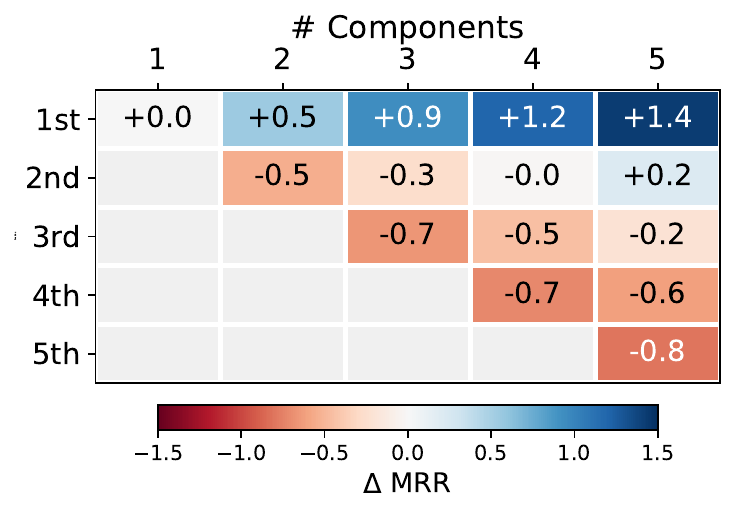}
\caption{\textbf{Position sensitivity of the $\tilde{a}$ component.} We report the change in MRR relative to the mean performance of queries sharing the same component set, grouped by the position of $\tilde{a}$ in the query.}
    \label{fig:position_heatmap}
\end{figure}

\paragraph{Component ordering matters for dense retrieval.}
Consistent with Section~\ref{sec:preliminary}, component order matters for dense retrieval.
In ANCE, performance is generally higher when $\tilde{a}$ appears earlier in the sequence, as shown in Figure~\ref{fig:position_heatmap}.
This may reflect the encoding behavior of dense retrievers, where earlier tokens can have a stronger influence on the final query representation.
Additional analyses on the number of queries are provided in Appendix~\ref{app:component_analysis}.
Overall, these results suggest that effective retrieval depends not only on query content, but also on structurally diverse query variations.

\section{Related Work}
In conversational search, the main challenge is to interpret the current query in light of the preceding dialogue context. To address this challenge, two main approaches have been explored: conversational dense retrieval (CDR) and conversational query reformulation (CQR).

CDR methods encode conversational context directly for retrieval without an explicit query rewriting step~\cite{cdr_chen, cdr_lin, cdr_mao}. Representative approaches include ConvDR, which learns contextualized dense embeddings via teacher--student distillation~\cite{cdr_yu}, and Contextualized Query Embeddings, which develop a compact retriever for low-latency conversational passage retrieval~\cite{cdr_lin}. More recent work such as ConvAug improves generalization through LLM-based conversation augmentation~\cite{cdr_chen}.
A parallel line of work adapts sparse retrievers to the conversational setting: Conv-SPLADE fine-tunes SPLADE~\cite{cdr_splade} on conversational data.

In contrast, CQR enables the use of off-the-shelf retrievers by reformulating the current query into a standalone form that can be directly processed by existing retrieval models~\cite{itercqr, llm4cs, convsearchr1}. Recent CQR research has increasingly been shaped by large language models (LLMs). IterCQR~\cite{itercqr} uses LLM-generated rewrites for initialization instead of human annotations. LLM4CS~\cite{llm4cs} applies direct prompting to generate multiple rewrites and hypothetical responses.
LLM-aided informative query rewriting~\cite{llm-aided} employs LLMs in a rewrite-then-edit framework. ConvSearch-R1~\cite{convsearchr1} further explores training LLM-based reformulation models through supervised fine-tuning and reinforcement learning. Our work shares this motivation and aims to fully leverage LLM capabilities for CQR while improving efficiency through compositional generation.
% for more robust query representations

%while CMQR~\cite{cmqr} combines multiple rewrites using beam-search scores for retrieval. 

\section{Conclusion}
In conclusion, we introduce CompCQR, a simple yet effective training-free CQR framework. Rather than relying on costly training or repeated LLM inference, our approach generates diverse compositional reformulations, aggregates retrieval results across candidates, and uses lightweight answerability-guided reasoning to select high-quality evidence. Our findings show that: (1) our framework outperforms existing training- and prompting-based methods without explicit training labels across both dense and sparse retrievers; (2) document-level aggregation via RRF increasingly outperforms query-level aggregation as the number of candidates grows; and (3) our framework remains robust in long conversations, where the performance gap over existing QR methods widens as the conversation progresses.

\section*{Limitations}
Our work has several limitations. The current evaluation is limited to English benchmarks, so it remains unclear how well the framework transfers to other languages. Extending the approach to multilingual settings would therefore be an important direction for future work. In addition, while we consider a predefined set of compositional queries, we do not study how to identify the most effective subset for a given dataset or retriever. A more principled composition selection strategy could further improve both efficiency and retrieval effectiveness.

Although our experiments show that the framework is already cost-effective in terms of both LLM calls and FLOPs, the current design still requires multiple retriever calls to explore compositional variants. Further improving retrieval efficiency, for example by reducing the number of retrieval calls while preserving coverage, could lower end-to-end latency and make the framework even more practical. Our current framework also considers only permutations without repetition. Exploring more flexible composition strategies, including those that intentionally repeat selected components, adding more future queries, could further enrich the space of query variants and potentially improve retrieval effectiveness. Finally, while we evaluate compact open-source and API-based models, it would also be interesting to examine how the framework performs with larger open-source LLMs, such as 70B-scale models.
\section*{Acknowledgments}
This work was partly supported by the Institute of Information \& Communications Technology Planning \& Evaluation (IITP) grant funded by the Korea government (MSIT) (No. RS-2022-II220184, Development and Study of AI Technologies to Inexpensively Conform to Evolving Policy on Ethics), and the Artificial Intelligence Graduate School Program (Seoul National University) (No. RS-2021-II212068)). K. Jung is with the Automation and Systems Research Institute (ASRI), Seoul National University, Seoul, Republic of Korea.

\bibliography{custom}

@article{pytrec_eval,
  title={Pytrec\_eval: An Extremely Fast Python Interface to trec\_eval},
  author={Christophe Van Gysel and M. de Rijke},
  journal={The 41st International ACM SIGIR Conference on Research \& Development in Information Retrieval},
  year={2018},
  url={https://api.semanticscholar.org/CorpusID:13691943}
}

@article{topiocqa,
    title = "{T}opi{OCQA}: Open-domain Conversational Question Answering with Topic Switching",
    author = "Adlakha, Vaibhav  and
      Dhuliawala, Shehzaad  and
      Suleman, Kaheer  and
      de Vries, Harm  and
      Reddy, Siva",
    editor = "Roark, Brian  and
      Nenkova, Ani",
    journal = "Transactions of the Association for Computational Linguistics",
    volume = "10",
    year = "2022",
    address = "Cambridge, MA",
    publisher = "MIT Press",
    url = "https://aclanthology.org/2022.tacl-1.27/",
    doi = "10.1162/tacl_a_00471",
    pages = "468--483"
}

@inproceedings{qrecc,
    title = "Open-Domain Question Answering Goes Conversational via Question Rewriting",
    author = "Anantha, Raviteja  and
      Vakulenko, Svitlana  and
      Tu, Zhucheng  and
      Longpre, Shayne  and
      Pulman, Stephen  and
      Chappidi, Srinivas",
    editor = "Toutanova, Kristina  and
      Rumshisky, Anna  and
      Zettlemoyer, Luke  and
      Hakkani-Tur, Dilek  and
      Beltagy, Iz  and
      Bethard, Steven  and
      Cotterell, Ryan  and
      Chakraborty, Tanmoy  and
      Zhou, Yichao",
    booktitle = "Proceedings of the 2021 Conference of the North American Chapter of the Association for Computational Linguistics: Human Language Technologies",
    month = jun,
    year = "2021",
    address = "Online",
    publisher = "Association for Computational Linguistics",
    url = "https://aclanthology.org/2021.naacl-main.44/",
    doi = "10.18653/v1/2021.naacl-main.44",
    pages = "520--534"
}

@article{cast19,
  author       = {Jeffrey Dalton and
                  Chenyan Xiong and
                  Jamie Callan},
  title        = {{TREC} CAsT 2019: The Conversational Assistance Track Overview},
  journal      = {CoRR},
  volume       = {abs/2003.13624},
  year         = {2020},
  url          = {https://arxiv.org/abs/2003.13624},
  eprinttype    = {arXiv},
  eprint       = {2003.13624},
  bibsource    = {dblp computer science bibliography, https://dblp.org}
}

@inproceedings{cast20,
  title={CAsT 2020: The Conversational Assistance Track Overview},
  author={Jeffrey Dalton and Chenyan Xiong and Jamie Callan},
  booktitle={Text Retrieval Conference},
  year={2020},
  url={https://api.semanticscholar.org/CorpusID:214735659}
}

@inproceedings{itercqr,
    title = "{I}ter{CQR}: Iterative Conversational Query Reformulation with Retrieval Guidance",
    author = "Jang, Yunah  and
      Lee, Kang-il  and
      Bae, Hyunkyung  and
      Lee, Hwanhee  and
      Jung, Kyomin",
    editor = "Duh, Kevin  and
      Gomez, Helena  and
      Bethard, Steven",
    booktitle = "Proceedings of the 2024 Conference of the North American Chapter of the Association for Computational Linguistics: Human Language Technologies (Volume 1: Long Papers)",
    month = jun,
    year = "2024",
    address = "Mexico City, Mexico",
    publisher = "Association for Computational Linguistics",
    url = "https://aclanthology.org/2024.naacl-long.449/",
    doi = "10.18653/v1/2024.naacl-long.449",
    pages = "8121--8138"
}

@inproceedings{convsearchr1, 
    title = "{C}onv{S}earch-R1: Enhancing Query Reformulation for Conversational Search with Reasoning via Reinforcement Learning",
    author = "Zhu, Changtai  and
      Wang, Siyin  and
      Feng, Ruijun  and
      Song, Kai  and
      Qiu, Xipeng",
    editor = "Christodoulopoulos, Christos  and
      Chakraborty, Tanmoy  and
      Rose, Carolyn  and
      Peng, Violet",
    booktitle = "Proceedings of the 2025 Conference on Empirical Methods in Natural Language Processing",
    month = nov,
    year = "2025",
    address = "Suzhou, China",
    publisher = "Association for Computational Linguistics",
    url = "https://aclanthology.org/2025.emnlp-main.1349/",
    doi = "10.18653/v1/2025.emnlp-main.1349",
    pages = "26547--26564",
    ISBN = "979-8-89176-332-6"
}

@inproceedings{adarewriter,
    title = "{A}da{R}ewriter: Unleashing the Power of Prompting-based Conversational Query Reformulation via Test-Time Adaptation",
    author = "Lai, Yilong  and
      Wu, Jialong  and
      Wang, Zhenglin  and
      Zhou, Deyu",
    editor = "Christodoulopoulos, Christos  and
      Chakraborty, Tanmoy  and
      Rose, Carolyn  and
      Peng, Violet",
    booktitle = "Proceedings of the 2025 Conference on Empirical Methods in Natural Language Processing",
    month = nov,
    year = "2025",
    address = "Suzhou, China",
    publisher = "Association for Computational Linguistics",
    url = "https://aclanthology.org/2025.emnlp-main.193/",
    doi = "10.18653/v1/2025.emnlp-main.193",
    pages = "3889--3905",
    ISBN = "979-8-89176-332-6"
}

@inproceedings{adacqr,
    title = "{A}da{CQR}: Enhancing Query Reformulation for Conversational Search via Sparse and Dense Retrieval Alignment",
    author = "Lai, Yilong  and
      Wu, Jialong  and
      Zhang, Congzhi  and
      Sun, Haowen  and
      Zhou, Deyu",
    editor = "Rambow, Owen  and
      Wanner, Leo  and
      Apidianaki, Marianna  and
      Al-Khalifa, Hend  and
      Eugenio, Barbara Di  and
      Schockaert, Steven",
    booktitle = "Proceedings of the 31st International Conference on Computational Linguistics",
    month = jan,
    year = "2025",
    address = "Abu Dhabi, UAE",
    publisher = "Association for Computational Linguistics",
    url = "https://aclanthology.org/2025.coling-main.515/",
    pages = "7698--7720"
}

@inproceedings{convgqr,
    title = "{C}onv{GQR}: Generative Query Reformulation for Conversational Search",
    author = "Mo, Fengran  and
      Mao, Kelong  and
      Zhu, Yutao  and
      Wu, Yihong  and
      Huang, Kaiyu  and
      Nie, Jian-Yun",
    editor = "Rogers, Anna  and
      Boyd-Graber, Jordan  and
      Okazaki, Naoaki",
    booktitle = "Proceedings of the 61st Annual Meeting of the Association for Computational Linguistics (Volume 1: Long Papers)",
    month = jul,
    year = "2023",
    address = "Toronto, Canada",
    publisher = "Association for Computational Linguistics",
    url = "https://aclanthology.org/2023.acl-long.274/",
    doi = "10.18653/v1/2023.acl-long.274",
    pages = "4998--5012"
}

@misc{agentic_conversational_search,
      title={Agentic Conversational Search with Contextualized Reasoning via Reinforcement Learning}, 
      author={Fengran Mo and Yifan Gao and Sha Li and Hansi Zeng and Xin Liu and Zhaoxuan Tan and Xian Li and Jianshu Chen and Dakuo Wang and Meng Jiang},
      year={2026},
      eprint={2601.13115},
      archivePrefix={arXiv},
      primaryClass={cs.CL},
      url={https://arxiv.org/abs/2601.13115}, 
}

@inproceedings{llm4cs,
    title = "Large Language Models Know Your Contextual Search Intent: A Prompting Framework for Conversational Search",
    author = "Mao, Kelong  and
      Dou, Zhicheng  and
      Mo, Fengran  and
      Hou, Jiewen  and
      Chen, Haonan  and
      Qian, Hongjin",
    editor = "Bouamor, Houda  and
      Pino, Juan  and
      Bali, Kalika",
    booktitle = "Findings of the Association for Computational Linguistics: EMNLP 2023",
    month = dec,
    year = "2023",
    address = "Singapore",
    publisher = "Association for Computational Linguistics",
    url = "https://aclanthology.org/2023.findings-emnlp.86/",
    doi = "10.18653/v1/2023.findings-emnlp.86",
    pages = "1211--1225"
}

@inproceedings{retpo,
    title = "Ask Optimal Questions: Aligning Large Language Models with Retriever{'}s Preference in Conversation",
    author = "Yoon, Chanwoong  and
      Kim, Gangwoo  and
      Jeon, Byeongguk  and
      Kim, Sungdong  and
      Jo, Yohan  and
      Kang, Jaewoo",
    editor = "Chiruzzo, Luis  and
      Ritter, Alan  and
      Wang, Lu",
    booktitle = "Findings of the Association for Computational Linguistics: NAACL 2025",
    month = apr,
    year = "2025",
    address = "Albuquerque, New Mexico",
    publisher = "Association for Computational Linguistics",
    url = "https://aclanthology.org/2025.findings-naacl.328/",
    doi = "10.18653/v1/2025.findings-naacl.328",
    pages = "5914--5936",
    ISBN = "979-8-89176-195-7"
}

@inproceedings{chiq,
    title = "{CHIQ}: Contextual History Enhancement for Improving Query Rewriting in Conversational Search",
    author = "Mo, Fengran  and
      Ghaddar, Abbas  and
      Mao, Kelong  and
      Rezagholizadeh, Mehdi  and
      Chen, Boxing  and
      Liu, Qun  and
      Nie, Jian-Yun",
    editor = "Al-Onaizan, Yaser  and
      Bansal, Mohit  and
      Chen, Yun-Nung",
    booktitle = "Proceedings of the 2024 Conference on Empirical Methods in Natural Language Processing",
    month = nov,
    year = "2024",
    address = "Miami, Florida, USA",
    publisher = "Association for Computational Linguistics",
    url = "https://aclanthology.org/2024.emnlp-main.135/",
    doi = "10.18653/v1/2024.emnlp-main.135",
    pages = "2253--2268"
}

@inproceedings{llm-aided,
    title = "Enhancing Conversational Search: Large Language Model-Aided Informative Query Rewriting",
    author = "Ye, Fanghua  and
      Fang, Meng  and
      Li, Shenghui  and
      Yilmaz, Emine",
    editor = "Bouamor, Houda  and
      Pino, Juan  and
      Bali, Kalika",
    booktitle = "Findings of the Association for Computational Linguistics: EMNLP 2023",
    month = dec,
    year = "2023",
    address = "Singapore",
    publisher = "Association for Computational Linguistics",
    url = "https://aclanthology.org/2023.findings-emnlp.398/",
    doi = "10.18653/v1/2023.findings-emnlp.398",
    pages = "5985--6006"
}

@misc{ance,
      title={Approximate Nearest Neighbor Negative Contrastive Learning for Dense Text Retrieval}, 
      author={Lee Xiong and Chenyan Xiong and Ye Li and Kwok-Fung Tang and Jialin Liu and Paul Bennett and Junaid Ahmed and Arnold Overwijk},
      year={2020},
      eprint={2007.00808},
      archivePrefix={arXiv},
      primaryClass={cs.IR},
      url={https://arxiv.org/abs/2007.00808}, 
}

@article{bm25,
author = {Robertson, Stephen and Zaragoza, Hugo},
year = {2009},
month = {01},
pages = {333-389},
title = {The Probabilistic Relevance Framework: BM25 and Beyond},
volume = {3},
journal = {Foundations and Trends in Information Retrieval},
doi = {10.1561/1500000019}
}

@article{pyserini,
  author       = {Jimmy Lin and
                  Xueguang Ma and
                  Sheng{-}Chieh Lin and
                  Jheng{-}Hong Yang and
                  Ronak Pradeep and
                  Rodrigo Nogueira},
  title        = {Pyserini: An Easy-to-Use Python Toolkit to Support Replicable {IR}
                  Research with Sparse and Dense Representations},
  journal      = {CoRR},
  volume       = {abs/2102.10073},
  year         = {2021},
  url          = {https://arxiv.org/abs/2102.10073},
  eprinttype    = {arXiv},
  eprint       = {2102.10073},
  bibsource    = {dblp computer science bibliography, https://dblp.org}
}

@misc{faiss,
      title={Billion-scale similarity search with GPUs}, 
      author={Jeff Johnson and Matthijs Douze and Hervé Jégou},
      year={2017},
      eprint={1702.08734},
      archivePrefix={arXiv},
      primaryClass={cs.CV},
      url={https://arxiv.org/abs/1702.08734}, 
}

@InProceedings{shapely1,
  title = 	 {Data Shapley: Equitable Valuation of Data for Machine Learning},
  author =       {Ghorbani, Amirata and Zou, James},
  booktitle = 	 {Proceedings of the 36th International Conference on Machine Learning},
  pages = 	 {2242--2251},
  year = 	 {2019},
  editor = 	 {Chaudhuri, Kamalika and Salakhutdinov, Ruslan},
  volume = 	 {97},
  series = 	 {Proceedings of Machine Learning Research},
  month = 	 {09--15 Jun},
  publisher =    {PMLR},
  url = 	 {https://proceedings.mlr.press/v97/ghorbani19c.html}
}

@inproceedings{shapely2,
  title     = {The Shapley Value in Machine Learning},
  author    = {Rozemberczki, Benedek and Watson, Lauren and Bayer, Péter and Yang, Hao-Tsung and Kiss, Olivér and Nilsson, Sebastian and Sarkar, Rik},
  booktitle = {Proceedings of the Thirty-First International Joint Conference on
               Artificial Intelligence, {IJCAI-22}},
  publisher = {International Joint Conferences on Artificial Intelligence Organization},
  editor    = {Lud De Raedt},
  pages     = {5572--5579},
  year      = {2022},
  month     = {7},
  note      = {Survey Track},
  doi       = {10.24963/ijcai.2022/778},
  url       = {https://doi.org/10.24963/ijcai.2022/778},
}

@article{survey,
author = {Mo, Fengran and Mao, Kelong and Zhao, Ziliang and Qian, Hongjin and Chen, Haonan and Cheng, Yiruo and Li, Xiaoxi and Zhu, Yutao and Dou, Zhicheng and Nie, Jian-Yun},
title = {A Survey of Conversational Search},
year = {2025},
issue_date = {November 2025},
publisher = {Association for Computing Machinery},
address = {New York, NY, USA},
volume = {43},
number = {6},
issn = {1046-8188},
url = {https://doi.org/10.1145/3759453},
doi = {10.1145/3759453},
journal = {ACM Trans. Inf. Syst.},
month = sep,
articleno = {167},
numpages = {50}
}

@inproceedings{searcho1,
    title = "Search-o1: Agentic Search-Enhanced Large Reasoning Models",
    author = "Li, Xiaoxi  and
      Dong, Guanting  and
      Jin, Jiajie  and
      Zhang, Yuyao  and
      Zhou, Yujia  and
      Zhu, Yutao  and
      Zhang, Peitian  and
      Dou, Zhicheng",
    editor = "Christodoulopoulos, Christos  and
      Chakraborty, Tanmoy  and
      Rose, Carolyn  and
      Peng, Violet",
    booktitle = "Proceedings of the 2025 Conference on Empirical Methods in Natural Language Processing",
    month = nov,
    year = "2025",
    address = "Suzhou, China",
    publisher = "Association for Computational Linguistics",
    url = "https://aclanthology.org/2025.emnlp-main.276/",
    doi = "10.18653/v1/2025.emnlp-main.276",
    pages = "5420--5438",
    ISBN = "979-8-89176-332-6"
}

@misc{websearch,
      title={From Web Search towards Agentic Deep Research: Incentivizing Search with Reasoning Agents}, 
      author={Weizhi Zhang and Yangning Li and Yuanchen Bei and Junyu Luo and Guancheng Wan and Liangwei Yang and Chenxuan Xie and Yuyao Yang and Wei-Chieh Huang and Chunyu Miao and Henry Peng Zou and Xiao Luo and Yusheng Zhao and Yankai Chen and Chunkit Chan and Peilin Zhou and Xinyang Zhang and Chenwei Zhang and Jingbo Shang and Ming Zhang and Yangqiu Song and Irwin King and Philip S. Yu},
      year={2025},
      eprint={2506.18959},
      archivePrefix={arXiv},
      primaryClass={cs.IR},
      url={https://arxiv.org/abs/2506.18959}, 
}

@inproceedings{cdr_chen,
    title = "Generalizing Conversational Dense Retrieval via {LLM}-Cognition Data Augmentation",
    author = "Chen, Haonan  and
      Dou, Zhicheng  and
      Mao, Kelong  and
      Liu, Jiongnan  and
      Zhao, Ziliang",
    editor = "Ku, Lun-Wei  and
      Martins, Andre  and
      Srikumar, Vivek",
    booktitle = "Proceedings of the 62nd Annual Meeting of the Association for Computational Linguistics (Volume 1: Long Papers)",
    month = aug,
    year = "2024",
    address = "Bangkok, Thailand",
    publisher = "Association for Computational Linguistics",
    url = "https://aclanthology.org/2024.acl-long.149/",
    doi = "10.18653/v1/2024.acl-long.149",
    pages = "2700--2718"
}

@inproceedings{rrf,
author = {Cormack, Gordon V. and Clarke, Charles L A and Buettcher, Stefan},
title = {Reciprocal rank fusion outperforms condorcet and individual rank learning methods},
year = {2009},
isbn = {9781605584836},
publisher = {Association for Computing Machinery},
address = {New York, NY, USA},
url = {https://doi.org/10.1145/1571941.1572114},
doi = {10.1145/1571941.1572114},
booktitle = {Proceedings of the 32nd International ACM SIGIR Conference on Research and Development in Information Retrieval},
pages = {758–759},
numpages = {2},
location = {Boston, MA, USA},
series = {SIGIR '09}
}

@inproceedings{cmqr,
author = {Kostric, Ivica and Balog, Krisztian},
title = {A Surprisingly Simple yet Effective Multi-Query Rewriting Method for Conversational Passage Retrieval},
year = {2024},
isbn = {9798400704314},
publisher = {Association for Computing Machinery},
address = {New York, NY, USA},
url = {https://doi.org/10.1145/3626772.3657933},
doi = {10.1145/3626772.3657933},
booktitle = {Proceedings of the 47th International ACM SIGIR Conference on Research and Development in Information Retrieval},
pages = {2271–2275},
numpages = {5},
location = {Washington DC, USA},
series = {SIGIR '24}
}

@inproceedings{
llm_lost,
title={{LLM}s Get Lost In Multi-Turn Conversation},
author={Philippe Laban and Hiroaki Hayashi and Yingbo Zhou and Jennifer Neville},
booktitle={The Fourteenth International Conference on Learning Representations},
year={2026},
url={https://openreview.net/forum?id=VKGTGGcwl6}
}

@article{multi-turn_llm,
  publtype={informal},
  author={Shengyue Guan and Haoyi Xiong and Jindong Wang and Jiang Bian and Bin Zhu and Jian-guang Lou},
  title={Evaluating LLM-based Agents for Multi-Turn Conversations: A Survey},
  year={2025},
  month={March},
  cdate={1740787200000},
  journal={CoRR},
  volume={abs/2503.22458},
  url={https://doi.org/10.48550/arXiv.2503.22458}
}

@inproceedings{
multi-turn,
title={{MINT}: Evaluating {LLM}s in Multi-turn Interaction with Tools and Language Feedback},
author={Xingyao Wang and Zihan Wang and Jiateng Liu and Yangyi Chen and Lifan Yuan and Hao Peng and Heng Ji},
booktitle={The Twelfth International Conference on Learning Representations},
year={2024},
url={https://openreview.net/forum?id=jp3gWrMuIZ}
}

@inproceedings{cqr,
author = {Yu, Shi and Liu, Jiahua and Yang, Jingqin and Xiong, Chenyan and Bennett, Paul and Gao, Jianfeng and Liu, Zhiyuan},
title = {Few-Shot Generative Conversational Query Rewriting},
year = {2020},
isbn = {9781450380164},
publisher = {Association for Computing Machinery},
address = {New York, NY, USA},
url = {https://doi.org/10.1145/3397271.3401323},
doi = {10.1145/3397271.3401323},
booktitle = {Proceedings of the 43rd International ACM SIGIR Conference on Research and Development in Information Retrieval},
pages = {1933–1936},
numpages = {4},
location = {Virtual Event, China},
series = {SIGIR '20}
}

@inproceedings{conqrr,
    title = "{CONQRR}: Conversational Query Rewriting for Retrieval with Reinforcement Learning",
    author = "Wu, Zeqiu  and
      Luan, Yi  and
      Rashkin, Hannah  and
      Reitter, David  and
      Hajishirzi, Hannaneh  and
      Ostendorf, Mari  and
      Tomar, Gaurav Singh",
    editor = "Goldberg, Yoav  and
      Kozareva, Zornitsa  and
      Zhang, Yue",
    booktitle = "Proceedings of the 2022 Conference on Empirical Methods in Natural Language Processing",
    month = dec,
    year = "2022",
    address = "Abu Dhabi, United Arab Emirates",
    publisher = "Association for Computational Linguistics",
    url = "https://aclanthology.org/2022.emnlp-main.679/",
    doi = "10.18653/v1/2022.emnlp-main.679",
    pages = "10000--10014"
}

@article{t5qr,
  author       = {Sheng{-}Chieh Lin and
                  Jheng{-}Hong Yang and
                  Rodrigo Nogueira and
                  Ming{-}Feng Tsai and
                  Chuan{-}Ju Wang and
                  Jimmy Lin},
  title        = {Conversational Question Reformulation via Sequence-to-Sequence Architectures
                  and Pretrained Language Models},
  journal      = {CoRR},
  volume       = {abs/2004.01909},
  year         = {2020},
  url          = {https://arxiv.org/abs/2004.01909},
  eprinttype    = {arXiv},
  eprint       = {2004.01909},
  bibsource    = {dblp computer science bibliography, https://dblp.org}
}

@inproceedings{edirics, 
    title = "Search-Oriented Conversational Query Editing",
    author = "Mao, Kelong  and
      Dou, Zhicheng  and
      Liu, Bang  and
      Qian, Hongjin  and
      Mo, Fengran  and
      Wu, Xiangli  and
      Cheng, Xiaohua  and
      Cao, Zhao",
    editor = "Rogers, Anna  and
      Boyd-Graber, Jordan  and
      Okazaki, Naoaki",
    booktitle = "Findings of the Association for Computational Linguistics: ACL 2023",
    month = jul,
    year = "2023",
    address = "Toronto, Canada",
    publisher = "Association for Computational Linguistics",
    url = "https://aclanthology.org/2023.findings-acl.256/",
    doi = "10.18653/v1/2023.findings-acl.256",
    pages = "4160--4172"
}

@inproceedings{cdr_lin,
    title = "Contextualized Query Embeddings for Conversational Search",
    author = "Lin, Sheng-Chieh  and
      Yang, Jheng-Hong  and
      Lin, Jimmy",
    editor = "Moens, Marie-Francine  and
      Huang, Xuanjing  and
      Specia, Lucia  and
      Yih, Scott Wen-tau",
    booktitle = "Proceedings of the 2021 Conference on Empirical Methods in Natural Language Processing",
    month = nov,
    year = "2021",
    address = "Online and Punta Cana, Dominican Republic",
    publisher = "Association for Computational Linguistics",
    url = "https://aclanthology.org/2021.emnlp-main.77/",
    doi = "10.18653/v1/2021.emnlp-main.77",
    pages = "1004--1015"
}

@inproceedings{cdr_mao,
author = {Mao, Kelong and Qian, Hongjin and Mo, Fengran and Dou, Zhicheng and Liu, Bang and Cheng, Xiaohua and Cao, Zhao},
title = {Learning Denoised and Interpretable Session Representation for Conversational Search},
year = {2023},
isbn = {9781450394161},
publisher = {Association for Computing Machinery},
address = {New York, NY, USA},
url = {https://doi.org/10.1145/3543507.3583265},
doi = {10.1145/3543507.3583265},
booktitle = {Proceedings of the ACM Web Conference 2023},
pages = {3193–3202},
numpages = {10},
location = {Austin, TX, USA},
series = {WWW '23}
}

@inproceedings{cdr-instructor,
    title = "{I}nstructo{R}: Instructing Unsupervised Conversational Dense Retrieval with Large Language Models",
    author = "Jin, Zhuoran  and
      Cao, Pengfei  and
      Chen, Yubo  and
      Liu, Kang  and
      Zhao, Jun",
    editor = "Bouamor, Houda  and
      Pino, Juan  and
      Bali, Kalika",
    booktitle = "Findings of the Association for Computational Linguistics: EMNLP 2023",
    month = dec,
    year = "2023",
    address = "Singapore",
    publisher = "Association for Computational Linguistics",
    url = "https://aclanthology.org/2023.findings-emnlp.443/",
    doi = "10.18653/v1/2023.findings-emnlp.443",
    pages = "6649--6675"
}

@inproceedings{cdr_splade,
author = {Formal, Thibault and Piwowarski, Benjamin and Clinchant, St\'{e}phane},
title = {SPLADE: Sparse Lexical and Expansion Model for First Stage Ranking},
year = {2021},
isbn = {9781450380379},
publisher = {Association for Computing Machinery},
address = {New York, NY, USA},
url = {https://doi.org/10.1145/3404835.3463098},
doi = {10.1145/3404835.3463098},
booktitle = {Proceedings of the 44th International ACM SIGIR Conference on Research and Development in Information Retrieval},
pages = {2288–2292},
numpages = {5},
location = {Virtual Event, Canada},
series = {SIGIR '21}
}

@inproceedings{cdr_yu,
author = {Yu, Shi and Liu, Zhenghao and Xiong, Chenyan and Feng, Tao and Liu, Zhiyuan},
title = {Few-Shot Conversational Dense Retrieval},
year = {2021},
isbn = {9781450380379},
publisher = {Association for Computing Machinery},
address = {New York, NY, USA},
url = {https://doi.org/10.1145/3404835.3462856},
doi = {10.1145/3404835.3462856},
booktitle = {Proceedings of the 44th International ACM SIGIR Conference on Research and Development in Information Retrieval},
pages = {829–838},
numpages = {10},
location = {Virtual Event, Canada},
series = {SIGIR '21}
}

@inproceedings{icr,
    title = "{ICR}: Iterative Clarification and Rewriting for Conversational Search",
    author = "Cao, Zhiyu  and
      Li, Peifeng  and
      Zhu, Qiaoming",
    editor = "Christodoulopoulos, Christos  and
      Chakraborty, Tanmoy  and
      Rose, Carolyn  and
      Peng, Violet",
    booktitle = "Proceedings of the 2025 Conference on Empirical Methods in Natural Language Processing",
    month = nov,
    year = "2025",
    address = "Suzhou, China",
    publisher = "Association for Computational Linguistics",
    url = "https://aclanthology.org/2025.emnlp-main.496/",
    doi = "10.18653/v1/2025.emnlp-main.496",
    pages = "9810--9824",
    ISBN = "979-8-89176-332-6"
}

@inproceedings{followup_query,
author = {Meng, Chuan and Tonolini, Francesco and Mo, Fengran and Aletras, Nikolaos and Yilmaz, Emine and Kazai, Gabriella},
title = {Bridging the Gap: From Ad-hoc to Proactive Search in Conversations},
year = {2025},
isbn = {9798400715921},
publisher = {Association for Computing Machinery},
address = {New York, NY, USA},
url = {https://doi.org/10.1145/3726302.3729915},
doi = {10.1145/3726302.3729915},
booktitle = {Proceedings of the 48th International ACM SIGIR Conference on Research and Development in Information Retrieval},
pages = {64–74},
numpages = {11},
location = {Padua, Italy},
series = {SIGIR '25}
}

@inproceedings{proactive_benchmark,
author = {Samarinas, Chris and Zamani, Hamed},
title = {ProCIS: A Benchmark for Proactive Retrieval in Conversations},
year = {2024},
isbn = {9798400704314},
publisher = {Association for Computing Machinery},
address = {New York, NY, USA},
url = {https://doi.org/10.1145/3626772.3657869},
doi = {10.1145/3626772.3657869},
booktitle = {Proceedings of the 47th International ACM SIGIR Conference on Research and Development in Information Retrieval},
pages = {830–840},
numpages = {11},
location = {Washington DC, USA},
series = {SIGIR '24}
}

@misc{genqrensemble,
      title={Generative Query Reformulation Using Ensemble Prompting, Document Fusion, and Relevance Feedback}, 
      author={Kaustubh D. Dhole and Ramraj Chandradevan and Eugene Agichtein},
      year={2024},
      eprint={2405.17658},
      archivePrefix={arXiv},
      primaryClass={cs.IR},
      url={https://arxiv.org/abs/2405.17658}, 
}

@inproceedings{cdr-chatretriever,
    title = "{C}hat{R}etriever: Adapting Large Language Models for Generalized and Robust Conversational Dense Retrieval",
    author = "Mao, Kelong  and
      Deng, Chenlong  and
      Chen, Haonan  and
      Mo, Fengran  and
      Liu, Zheng  and
      Sakai, Tetsuya  and
      Dou, Zhicheng",
    editor = "Al-Onaizan, Yaser  and
      Bansal, Mohit  and
      Chen, Yun-Nung",
    booktitle = "Proceedings of the 2024 Conference on Empirical Methods in Natural Language Processing",
    month = nov,
    year = "2024",
    address = "Miami, Florida, USA",
    publisher = "Association for Computational Linguistics",
    url = "https://aclanthology.org/2024.emnlp-main.71/",
    doi = "10.18653/v1/2024.emnlp-main.71",
    pages = "1227--1240"
}

@misc{cdr-convmix,
      title={ConvMix: A Mixed-Criteria Data Augmentation Framework for Conversational Dense Retrieval}, 
      author={Fengran Mo and Jinghan Zhang and Yuchen Hui and Jia Ao Sun and Zhichao Xu and Zhan Su and Jian-Yun Nie},
      year={2025},
      eprint={2508.04001},
      archivePrefix={arXiv},
      primaryClass={cs.IR},
      url={https://arxiv.org/abs/2508.04001}, 
}

@article{grattafiori2024llama,
  title={The llama 3 herd of models},
  author={Grattafiori, Aaron and Dubey, Abhimanyu and Jauhri, Abhinav and Pandey, Abhinav and Kadian, Abhishek and Al-Dahle, Ahmad and Letman, Aiesha and Mathur, Akhil and Schelten, Alan and Vaughan, Alex and others},
  journal={arXiv preprint arXiv:2407.21783},
  year={2024}
}

@misc{gpt4.1,
  author       = {OpenAI},
  title        = {Introducing {GPT-4.1} in the {API}},
  year         = {2025},
  howpublished = {\url{https://openai.com/index/gpt-4-1/}},
}

@inproceedings{gtr,
    title = "Large Dual Encoders Are Generalizable Retrievers",
    author = "Ni, Jianmo  and
      Qu, Chen  and
      Lu, Jing  and
      Dai, Zhuyun  and
      Hernandez Abrego, Gustavo  and
      Ma, Ji  and
      Zhao, Vincent  and
      Luan, Yi  and
      Hall, Keith  and
      Chang, Ming-Wei  and
      Yang, Yinfei",
    editor = "Goldberg, Yoav  and
      Kozareva, Zornitsa  and
      Zhang, Yue",
    booktitle = "Proceedings of the 2022 Conference on Empirical Methods in Natural Language Processing",
    month = dec,
    year = "2022",
    address = "Abu Dhabi, United Arab Emirates",
    publisher = "Association for Computational Linguistics",
    url = "https://aclanthology.org/2022.emnlp-main.669/",
    doi = "10.18653/v1/2022.emnlp-main.669",
    pages = "9844--9855"
}

@article{contriever,
  author       = {Gautier Izacard and
                  Mathilde Caron and
                  Lucas Hosseini and
                  Sebastian Riedel and
                  Piotr Bojanowski and
                  Armand Joulin and
                  Edouard Grave},
  title        = {Towards Unsupervised Dense Information Retrieval with Contrastive
                  Learning},
  journal      = {CoRR},
  volume       = {abs/2112.09118},
  year         = {2021},
  url          = {https://arxiv.org/abs/2112.09118},
  eprinttype   = {arXiv},
  eprint       = {2112.09118},
  bibsource    = {dblp computer science bibliography, https://dblp.org}
}

@misc{howyouaskmatters,
      title={How You Ask Matters! Adaptive RAG Robustness to Query Variations}, 
      author={Yunah Jang and Megha Sundriyal and Kyomin Jung and Meeyoung Cha},
      year={2026},
      eprint={2604.10745},
      archivePrefix={arXiv},
      primaryClass={cs.CL},
      url={https://arxiv.org/abs/2604.10745}, 
}
\clearpage
\appendix

\label{sec:appendix}

% \begin{table}[t]
% \centering
% \small
% \begin{threeparttable}

% \begin{tabular}{lcccccc}
% \toprule
% \multirow{2}{*}{Setting}
% & \multicolumn{3}{c}{CAsT19}
% & \multicolumn{3}{c}{CAsT20} \\
% \cmidrule(lr){2-4} \cmidrule(lr){5-7}
% & MRR & NDCG@3 & R@10
% & MRR & NDCG@3 & R@100 \\
% \midrule

% Ours (sorting, LLM, rule selection) &  &  &  &  &  &  \\
% - LLM selection (sorting, rule selection) &  &  &  &  &  &  \\
% - rule selection (sorting, LLM) &  &  &  &  &  &  \\
% - rule selection, LLM (sorting, doc list as run) &  &  &  &  &  &  \\

% \bottomrule
% \end{tabular}

% \caption{\textbf{Ablation study on CAsT19 and CAsT20}}
% \end{threeparttable}
% \end{table}

\section{Experimental Setting}

\subsection{Implementation Details}
\label{sec:appendix_implementationdetails}
All experiments are conducted using two NVIDIA A6000 GPUs. We implement the retrieval systems using Faiss~\cite{faiss} and Pyserini~\cite{pyserini}. For BM25, we follow previous work~\cite{itercqr, adacqr, adarewriter}. Specifically, we use $k_1 = 0.9$ and $b = 0.4$ for TopiOCQA, and $k_1 = 0.82$ and $b = 0.68$ for QReCC. 

For dense retrieval, we use ANCE~\cite{ance}, trained on MSMARCO, with maximum input lengths of 128 tokens for reformulated queries and 384 tokens for passages. Additionally, we report the results on GTR \cite{gtr}, and Contriever~\cite{contriever} on TopiOCQA. 

Because the best document selection step may vary across runs due to LLM sampling, we additionally report multi-run statistics for this component in Appendix~\ref{appendix:multi-run}, while the main results are reported from a single run.
For all comparisons with LLM4CS~\cite{llm4cs}, we implement the best-performing setting, Rewrite-and-Response (RAR), Mean Aggregation, and Chain-of-Thought (CoT). Since LLM4CS aggregates its generated queries by averaging query embeddings, it is not directly applicable to BM25. For sparse retrieval we therefore concatenate the rewrite and its response into a single query.

\begin{figure}[h!]
    \centering
    \includegraphics[width=\columnwidth]{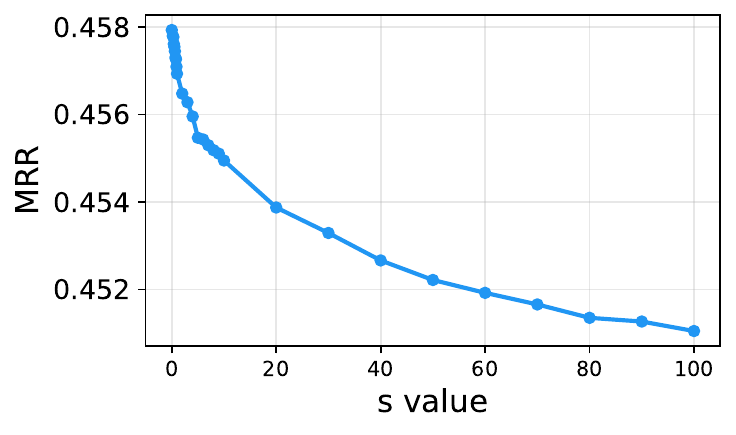}
    \vspace{-2mm}
    \caption{\textbf{RRF Softening $s$.} We test the impact of softening parameter s on RRF ranking.}
    \label{fig:rrf_k_value}
\end{figure}

\subsection{Hyperparameters}
\label{app:hyperparameters}

\subsubsection{RRF Softening Hyperparameter}
For the RRF score softening hyperparameter \(s\), we tested values from 0 to 100 on the validation set.
As shown in Figure~\ref{fig:rrf_k_value}, varying the \(s\) value does not significantly affect the final MRR score, although \(s=0\) achieves the best performance.
All results are therefore reported using \(s=0\).

\subsubsection{LLM Document Selection}
% \begin{table}[t]
% \centering
% \small
% \resizebox{\linewidth}{!}{
% \begin{tabular}{lcccc}
% \toprule[1.2pt]
% Model & K & Accuracy & MRR & R@5 \\
% \midrule

% \multirow{7}{*}{GPT-4.1-mini} 
% & 3 & 72.2\% & 44.3 & 55.5\\
% & 5 & 65.4\% & 45.8 & \\
% & 10 & 59.2\% & 47.6 & 57.6\\
% & 15 & 56.0\% & 48.0 & 58.4\\
% & 20 & 54.4\% & 48.5 & 58.9\\
% & 30 & 51.9\% & 48.6 & 59.0\\
% & 40 & 50.1\% & 48.7 & 59.6\\
% \midrule
% \multirow{7}{*}{Llama-3.1-8B} 
% & 3 & 61.6\% & 41.3& \\
% & 5 & 52.9\% & 41.3 &\\
% & 10 & 45.2\% & 41.3 &\\
% & 15 & 42.4\% & 41.3 &\\
% & 20 & 40.9\% & 41.5 &\\
% & 30 & 39.3\%  & 41.8 &\\
% %& 40 &  & &\\
% \bottomrule[1.2pt]
% \end{tabular}
% }
% \caption{\textbf{LLM best document selection accuracy.}
% We report the accuracy of LLMs on best document selection.}
% \label{tab:best-doc-accuracy}
% \end{table}

\begin{table}[t]
\centering
\small
\resizebox{\linewidth}{!}{
\begin{tabular}{lcccc}
\toprule[1.2pt]
Model & K & Accuracy & MRR & R@5 \\
\midrule

\multirow{6}{*}{GPT-4.1-mini} 
& 5 & 66.4\% & 41.5 & 49.7 \\
& 10 & 56.1\% & 42.0 & 52.3\\
& 20 & 53.2\% & 44.3 & 54.6\\
& 30 & 49.9\% & 44.6 & 54.0 \\
& 40 & 49.4\% & 45.1 & 54.6 \\
& 50 & 47.2\% & 44.7 & 54.7 \\
\midrule
\multirow{3}{*}{Llama-3.1-8B} 
& 5 & 54.0\% &  37.7 & 49.7\\
& 10 & 49.5\% & 39.7 & 51.0\\
& 20 & 39.9\% &  38.5  & 50.4\\
\bottomrule[1.2pt]
\end{tabular}
}
\caption{\textbf{LLM best document selection accuracy.}
We report the accuracy of LLMs on best document selection.}
\label{tab:best-doc-accuracy}
\end{table}
\begin{table}[t]
\centering
\small
\resizebox{\linewidth}{!}{
\begin{tabular}{lcccc}
\toprule[1.3pt]
QR Model & Selection Model & Accuracy & MRR \\
\midrule
GPT-4.1-mini  & Llama-3.1-8B & 48.3\%  & 43.7 (\textcolor{red}{\scriptsize -5.0}) \\
Llama-3.1-8B  & GPT-4.1-mini & 52.1\% & 46.4 (\textcolor{blue}{\scriptsize +4.1}) \\
\bottomrule[1.3pt]
\end{tabular}
}
\caption{\textbf{Impact of selection models.} We analyze the effects of the QR model and the document selection model by using different models for each component. The $+$ and $-$ indicate the performance relative to using the original QR model as the selection model.}
\label{tab:doc_analysis}
\end{table}

In our framework, we use LLM reasoning to select the document that best answers a given question conditioned on its conversational history.
To study the effect of the candidate set size, we vary the number of top-$k$ retrieved documents provided to the LLM from 5 to 50 (20 for Llama).
We evaluate this step using two metrics: (1) best-document accuracy, which measures how often the LLM selects the best document when it is included in the top-$k$ candidates, and (2) the final retrieval performance measured by MRR.

Table~\ref{tab:best-doc-accuracy} shows that the best-document selection accuracy of both models decreases as the number of candidate documents increases.
This trend suggests that document selection becomes more challenging as the LLM must reason over a larger candidate set.
However, for GPT-4.1-mini, the final retrieval performance continues to improve as $k$ increases, with MRR reaching its best value at $k=40$.
In contrast, Llama-3.1-8B shows substantially lower selection accuracy overall, and its performance becomes unstable when the number of candidate documents exceeds 20.
This indicates that Llama-3.1-8B struggles to follow the selection instructions reliably in larger candidate sets.

To further understand whether the lower performance of Llama-3.1-8B comes from weaker candidate generation or from the document selection step itself, we conduct an additional analysis in Table~\ref{tab:doc_analysis}.
When GPT-4.1-mini is replaced with Llama-3.1-8B as the selection model, the MRR drops by 5.0 points.
Conversely, when GPT-4.1-mini is used as the selection model in the Llama-based setting, the MRR improves by 4.1 points.
These results indicate that the final performance is influenced by both the quality of the query candidates and the effectiveness of the best-document selection model.

Based on the validation results, we set the number of input documents in the document selection stage according to each model's capacity.
Specifically, GPT-4.1-mini performs best with the top 40 documents on MRR, whereas Llama-3.1-8B remains most stable when given the top 10 documents.

\subsection{Dataset}
\begin{table}[t!]
\centering
\resizebox{\columnwidth}{!}{
\begin{tabular}{lcccc}
\toprule[1.4pt]
 & QReCC & TopiOCQA & CAsT-19 & CAsT-20 \\
\midrule
\# Dialogues   & 2775 & 205  & 50  & 25  \\
\# Turns       & 8209 & 2514 & 479 & 208 \\
\# Collections & 54M  & 25M  & 38M & 38M \\
\bottomrule[1.4pt]
\end{tabular}}
\caption{Statistics of conversational search test datasets: QReCC, TopiOCQA, CAsT-19, and CAsT-20.}
\label{tab:statistics}

\end{table}
In Table~\ref{tab:statistics}, we summarize the statistics of the datasets used in our experiments. All four datasets provide test sets with gold document annotations for conversational queries. The reported numbers correspond to the test instances with valid gold documents. Additionally, for hyperparameter tuning, we randomly sample 700 turns from the TopiOCQA training set.

\label{sec:statistics}

\section{Analysis of LLM-based Document Selection}

\subsection{Document Selection on Different Method}
\begin{table}[t]
\centering
\small
\begin{threeparttable}

\begin{adjustbox}{width=\linewidth}
\begin{tabular}{lcc}
\toprule[1.2pt]
Setting & MRR & NDCG@3 \\
\midrule

CompCQR (w/o LS) & 37.8 & 36.5  \\
CompCQR (ours - w/ LS) & \textbf{42.3} (+4.5) & \textbf{41.5} (+5.0) \\
\midrule
LLM4CS &35.2 & 34.8  \\
LLM4CS (w/ LS) & 36.4 (+1.2) & 36.7 (+1.9) \\
\bottomrule[1.2pt]
\end{tabular}
\end{adjustbox}

\caption{
\textbf{LLM4CS with top-1 document selection.}
We report the results obtained by applying top-1 LLM document selection (LS) to the LLM4CS method.
}
\label{tab:LLM4CS_top1doc}
\end{threeparttable}
\end{table}
\label{app:ls}
In this section, we further compare the performance of the baseline LLM4CS~\cite{llm4cs} with and without LLM-based best-document selection. We evaluate both settings on the TopiOCQA dataset using Llama-3.1-8B-Instruct. In Table~\ref{tab:LLM4CS_top1doc}, consistent with the results in Section~\ref{sec:aggregation_method}, our method benefits substantially from LLM-based document selection, achieving gains of up to 4.5 MRR points, whereas LLM4CS shows only a small increase of 1.2 points.

\subsection{Multi-run Statistics for Best-document Selection}
\label{appendix:multi-run}

To examine the stability of LLM-based document selection, we report multi-run statistics for the top-10 document selection setting. We focus on this setting because the stability of LLM decisions directly affects retrieval performance.

We run the LLM-based best-document selection process three times under the same top-10 setting and report the mean and standard deviation for each metric. The results show consistent performance across runs, with an MRR of 42.2 $\pm$ 0.9 and nDCG of 41.3 $\pm$ 0.6. These results suggest that LLM-based selection in the top-10 setting is reasonably stable and that our main findings are unlikely to be driven by run-to-run randomness.

\begin{table}[t]
\centering
\begin{adjustbox}{width=\linewidth}
\begin{tabular}{lccc}
\toprule[1.4pt]
 & MRR & NDCG@3 & R@5 \\
\midrule
RRF Aggregation & 37.8 & 36.5 & -  \\
+ w/ LS (ours) & \textbf{42.3} &  \textbf{41.5} &  \textbf{53.7} \\
+ w/ LR & 26.7 & 24.1 & 43.7 \\
\bottomrule[1.4pt]
\end{tabular}
\end{adjustbox}
\caption{\textbf{Performance comparison of LLM-based document selection and ranking.}
We compare RRF document aggregation with variants using LLM document selection (LS) and LLM document full reranking (LR).}
\label{tab:llm_reranking}
\end{table}
\subsection{Comparison with Re-ranking}
\label{app:llm_reranking}
We compare our top-1 document selection step with LLM-based reranking over the top 10 retrieved documents. We conduct the experiment on the TopiOCQA dataset using Llama-3.1-8B.

As shown in Table~\ref{tab:llm_reranking}, the results indicate that selecting a single top document is an easier task for the model, allowing it to identify the gold document more effectively. In contrast, reranking multiple documents is a more complex decision-making task that introduces additional noise, leading to substantially degraded performance.

\section{Component Analysis}
\label{app:component_analysis}

\subsection{Number of Queries and Components}

\begin{figure}[t]
    \centering
    \includegraphics[width=\columnwidth]{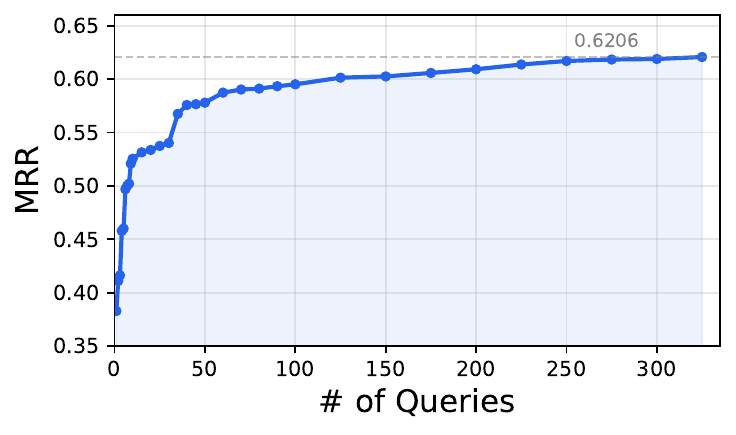}
    \caption{\textbf{Upper-bound MRR with accumulated queries.} We report upperbound of MRR on TopiOCQA.}
    \label{fig:rule_accumulation}
\end{figure}

Figure~\ref{fig:rule_accumulation} shows the upper-bound MRR on TopiOCQA with ANCE as the number of applied queries increases.
Retrieval performance improves steadily as more candidates are included, but the gains begin to saturate around 250 queries.
This suggests that broad compositional coverage is beneficial, while also indicating that selecting an optimal subset of queries could be a promising direction.

\section{Generalization to Different Models}
\subsection{Dense Retrievers}
\begin{table}[t]
\centering
\small
\begin{tabular}{lc}
\toprule
\textbf{Retriever} & \textbf{Top-10 Jaccard Distance} \\
\midrule
ANCE       & 28.0\% \\
GTR        & 22.6\% \\
Contriever & 23.2\% \\
\bottomrule
\end{tabular}
\caption{\textbf{Top-10 Jaccard Distance} under query-component reordering.}
\label{tab:additional_order_sensitivity}
\end{table}

\label{sec:different_retrievers}

We also report  the order sensitivity analysis from Section~\ref{sec:preliminary} using GTR and Contriever in Table~\ref{tab:additional_order_sensitivity}. Although the change rates for GTR and Contriever are lower than that of ANCE, both retrievers still exhibit substantial changes in their top-10 retrieved documents under reordering components. This suggests that order sensitivity is a general property of dense retrievers rather than an artifact specific to ANCE.

\subsection{Large Open-source Models}
\begin{table}[t]
\centering
\small
\resizebox{\columnwidth}{!}{
\begin{tabular}{llccc}
\toprule [1.2pt]
\textbf{Method} & \textbf{LLM} & \textbf{MRR} & \textbf{NDCG} & \textbf{R@100} \\
\midrule
LLM4CS & Qwen2.5-72B  & 43.4 & 42.2 & 82.2 \\
CompCQR  & Qwen2.5-72B  & \textbf{47.4} & \textbf{46.6} & \textbf{83.4} \\
\midrule
LLM4CS & Llama-3.3-70B & 45.3 & 44.5 & 83.6 \\
CompCQR  & Llama-3.3-70B & \textbf{48.6} & \textbf{47.5} & \textbf{84.0} \\
\bottomrule [1.2pt]
\end{tabular}}
\caption{\textbf{Retrieval performance with large open-source LLMs.} We report results on TopiOCQA using the ANCE retriever.}
\vspace{-1mm}
\label{tab:large_open_source_models}
\end{table}

We additionally evaluate CompCQR with two strong open-source LLMs, Qwen2.5-72B and Llama-3.3-70B, on TopiOCQA dataset with ANCE retriever.
As shown in Table~\ref{tab:large_open_source_models}, CompCQR consistently outperforms LLM4CS-5 across all retrieval metrics for both models. 
With Qwen2.5-72B, CompCQR improves MRR from 43.4 to 47.4, corresponding to a 9.2\% relative improvement. 
Similarly, with Llama-3.3-70B, CompCQR improves MRR from 45.3 to 48.6 (+7.3\% relative), with consistent gains in nDCG@3 and R@100. 
These results demonstrate that the effectiveness of CompCQR is not limited to a specific proprietary LLM and generalizes well to strong open-source LLMs.

\section{Topic Shift Performance}
\begin{figure}[t]
    \centering
    \includegraphics[width=\columnwidth]{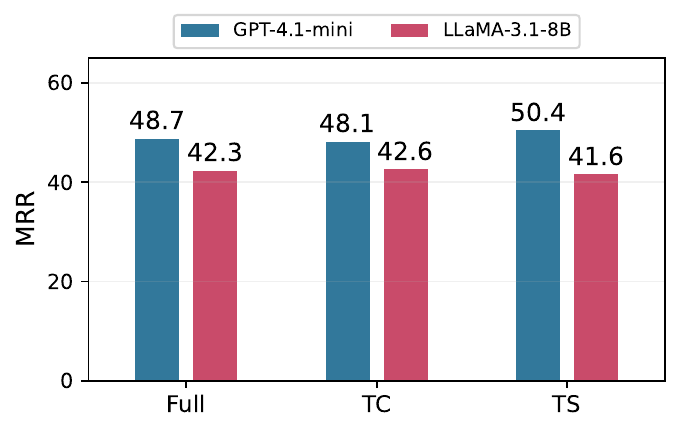}
    \vspace{-5mm}
    \caption{\textbf{Topic Shift Performance.} We test the framework performance in topic-shift (TS) and topic-concentrated (TC) turns.}
    \vspace{-1mm}
    \label{fig:topic_shift}
\end{figure}

We conduct an additional analysis on TopiOCQA by dividing the test instances according to whether the topic changes from the previous turn. 
Using the topic labels provided in the dataset, we define a turn as topic-shift (TS) when its topic differs from that of the preceding turn, and as topic-concentrated (TC) otherwise.

In Figure~\ref{fig:topic_shift}, results show a clear difference across models. GPT-4.1-mini preserves its performance under topic shifts and even achieves better results in those cases, while LLaMA-3.1-8B suffers from a slight degradation.
This discrepancy is likely attributable to differences in the underlying model capacity. 
More specifically, the findings indicate that the ability to correctly interpret conversational context plays a crucial role in handling topic transitions effectively.

\section{LLM Calls and FLOPs}
\label{app:computational_cost}
\begin{table}[t]
\centering
\small
\resizebox{\linewidth}{!}{
\begin{tabular}{lcccc}
\toprule[1.2pt]
Method & LLM Calls & TFLOPs & MRR \\
\midrule
CHIQ-Fusion &  6 & - & 38.0 \\
LLM4CS (k=5) & 5 &  \textbf{69} & 34.6 \\
LLM4CS (k=16) & 16 & 114 & 35.4 \\
\midrule
CompCQR (w/o LS) & \textbf{2} & 74 & 37.8 \\
CompCQR &3 & 109 & \textbf{42.3}\\

\bottomrule[1.2pt]
\end{tabular}
}
\caption{\textbf{Computational Cost.}
We compare the number of LLM calls, FLOPs per query, and corresponding TopiOCQA MRR across different methods. We report both CompCQR without LLM selection and the full framework. }
\label{tab:cost}
\end{table}
We compare the cost of CompCQR with prompting-based baselines in terms of LLM calls and FLOPs.
As shown in Table~\ref{tab:cost}, CompCQR achieves the best MRR with only three LLM calls, and the variant without LLM selection already surpasses LLM4CS ($k$=16) with two calls and fewer FLOPs.
This indicates that the gain comes from compositional generation rather than from additional inference.
We further note that although our framework processes more input tokens by conditioning on top-ranked documents during selection, this cost stems from longer input processing rather than repeated generation, which is typically cheaper in both latency and API pricing.

\section{Case Study}
\subsection{Mitigating Hard Negatives}
\begin{table*}[t!]
\centering
\small
\setlength{\tabcolsep}{6pt}
\renewcommand{\arraystretch}{1.12}

\begin{tabular}{
  >{\centering\arraybackslash}m{0.12\textwidth}
  >{\raggedright\arraybackslash}m{0.36\textwidth}
  >{\raggedright\arraybackslash}m{0.42\textwidth}
}
\specialrule{1.2pt}{1.5pt}{1.5pt}
\rowcolor{lightblue}
\multicolumn{3}{c}{
  \textbf{Original Query:} Describe core, aggregator, and edge switches.
} \\
\specialrule{1.2pt}{1.5pt}{2pt}
\textbf{Category} & \textbf{Document} & \textbf{LLM Passage Analysis} \\
\midrule

\textbf{RRF 1st-ranked}
&
``This makes a large \textbf{core switch} a logical way to handle traffic passing between ...
Top-of-rack switches are designed slightly differently from user \textbf{edge switches} ...''
&
\textbf{Passage 1} describes core switches and their roles, including their switching capacity and position in the network. ... with some details on data center use.
\\

\midrule

\textbf{LLM-selected (Rank: 3)}
&
``... an additional layer of switching, called the distribution layer, \textbf{aggregates the edge switches}. ...
The \textbf{aggregation layer} connects to the uplinks of the \textbf{edge switches} and is uplinked to the ... \textbf{network core}.''
&
\textbf{Passage 3} explicitly describes the distribution and aggregation layers, explaining how aggregation layers connect edge switches and aggregate uplinks. ... including the rationale behind the layers and design considerations.
\\

\midrule

\textbf{Reason for Selection}
& \multicolumn{2}{m{0.82\textwidth}}{
Passage 3 provides a clear and direct explanation of core, aggregator (aggregation), and edge switches ... making it the most suitable to answer the query comprehensively.}
\\

\bottomrule[1.3pt]
\end{tabular}

\caption{
\textbf{Case study of a hard negative in RRF top-1 ranking.}
Although the top-ranked RRF document is a hard negative, the LLM correctly identifies the relevant gold document that was originally ranked third.
}
\label{tab:case_hardneg_vs_llm_gold}
\end{table*}

In Table~\ref{tab:case_hardneg_vs_llm_gold}, we present a case in which LLM document selection step successfully distinguishes a hard negative from retrieved documents. The top-ranked document under RRF is topically relevant but only partially addresses the query, whereas the document selected by the LLM provides a more complete explanation. This shows that LLM-based selection can recover a suitable document even when rank fusion places a hard negative at the top.

\subsection{Failure Analysis}
\begin{table*}[t!]
\centering
\small
\renewcommand{\arraystretch}{1.35}
\begin{tabular}{lp{0.68\linewidth}c}
\toprule[1.3pt]
\textbf{Component} & \textbf{Content} & \textbf{Rank}\\
\midrule
Original Query &
Regarding the ancient kingdom above; was any innovation made that time? &
6 \\
Rewrite &
What innovations were made during the time of the ancient kingdom of Macedon? &
\textbf{1} \\
Rewrite Answer &
\ldots One notable innovation was the development of the phalanx formation, a tactical \ldots &
36 \\
Follow-up Query &
What were some of the notable innovations and achievements \ldots particularly during the reign of Alexander the Great? &
7 \\
Follow-up Answer &
\ldots particularly during the reign of Alexander the Great, was known for several notable innovations and achievements \ldots &
$>100$ \\
\midrule
CompCQR (RRF only) &
 &
8 \\
\bottomrule[1.3pt]
\end{tabular}
\caption{\textbf{A failure case from TopiOCQA.} We report the gold-passage rank for each component in the rightmost column. The generated components are truncated for readability.}
\label{tab:failure_case}
\end{table*}

We provide a concrete failure case from TopiOCQA where CompCQR underperforms a single-rewrite method. Table~\ref{tab:failure_case} shows the generated components and their corresponding gold-passage ranks.

In this example, the rewritten query retrieves the gold passage at rank 1. However, the auxiliary components introduce sub-topic drift: the rewrite answer focuses on military innovations such as the phalanx, while the follow-up components shift toward Alexander the Great's broader achievements. Since the gold passage instead concerns agricultural and glass-related innovations, these partially off-topic signals dilute the strong rewrite signal under reciprocal rank fusion, moving the gold passage from rank 1 to rank 8. This suggests that while component diversification is beneficial on average, it can hurt when generated components retrieve plausible but mismatched evidence.

\section{Comparison with CDR}
\begin{table}[t]
\centering
\resizebox{\columnwidth}{!}{
\begin{tabular}{lcccc}
\toprule[1.3pt]
Method & TopiOCQA & QReCC \\
\midrule
Conv-ANCE \cite{ance}              & 20.5 & 45.6  \\
ConvDR \cite{cdr_yu}                    & 26.4 & 35.7 \\
Conv-SPLADE \cite{cdr_splade}           & 29.5 & 46.6\\
InstructOR-ANCE \cite{cdr-instructor}          & 23.7 & 40.5\\
LeCoRE \cite{cdr_mao}                  & 32.0 & 51.1 \\
ConvAug \cite{cdr_chen}                 & 33.3 & 50.4\\
ChatRetriever \cite{cdr-chatretriever}     & 40.1 & 52.5 \\
ConvMix \cite{cdr-convmix} & 36.6 & 50.7 \\
\midrule
CompCQR (ours)  & \textbf{47.7}& \textbf{58.6} \\
\bottomrule[1.3pt]
\end{tabular}}
\caption{\textbf{Comparison with CDR Methods.} We report results on TopiOCQA, and QReCC with NDCG@3.}
\label{tab:cdr}
\end{table}
In this section, we compare CompCQR with various CDR methods. As shown in Table~\ref{tab:cdr}, our method achieves the best performance, outperforming the second-best CDR method by 7.6 NDCG@3 points on TopiOCQA. We use the same dense retriever as the Conv-ANCE (\citeyear{ance})  baseline.

\section{Baseline details.}
\label{sec:baseline_details}
We briefly summarize the baselines used in our experiments below.

\begingroup
\setlength{\itemsep}{0pt}
\setlength{\parskip}{0pt}
\setlength{\parsep}{0pt}
\begin{itemize}
    \item \textbf{T5QR}~\cite{t5qr}: A supervised T5-based query rewriting model fine-tuned to generate standalone rewrites for conversational search.
    \item \textbf{CONQRR}~\cite{conqrr}: A query rewriting model trained with reinforcement learning to directly optimize downstream retrieval performance for a fixed black-box retriever.
    \item \textbf{ConvGQR}~\cite{convgqr}: A trained generative reformulation framework that combines query rewriting with generated answer information for query expansion.
    \item \textbf{EdiRCS}~\cite{edirics}: A supervised text-editing-based conversational query rewriting model that copies most rewrite tokens from the dialogue context and generates only a small number of additional tokens.
    \item \textbf{IterCQR}~\cite{itercqr}: An iterative conversational query reformulation framework that optimizes the reformulation model using retrieval signals rather than relying solely on human rewrite annotations.
    \item \textbf{ReTPO}~\cite{retpo}: A retriever preference optimization framework that fine-tunes a query rewriting language model using retrieval feedback so that generated rewrites better align with the preferences of the target retriever.
    \item \textbf{ICR}~\cite{icr}: A trained iterative clarification-and-rewriting framework that improves reformulation quality by generating clarification signals before rewriting ambiguous conversational queries.
    \item \textbf{ConvSearch-R1}~\cite{convsearchr1}: A conversational reformulation framework trained with reinforcement learning to optimize query rewriting directly with retrieval-based rewards, without requiring rewrite supervision.
    \item \textbf{AdaCQR}~\cite{adacqr}: A hybrid reformulation framework that combines a fine-tuned query rewriting model with LLM-based fusion to improve robustness across retrieval settings.
    \item \textbf{CHIQ-Fusion}~\cite{chiq}: A hybrid method that combines a trained query rewriting model with LLM-based history enhancement and ambiguity resolution.
    \item \textbf{AdaRewriter}~\cite{adarewriter}: A prompting-based reformulation framework that uses test-time adaptation and a lightweight reward model to select the most effective rewrite candidate.
    \item \textbf{LLM-Aided}~\cite{llm-aided}: A training-free query rewriting approach that uses an LLM to generate more informative rewrites without task-specific model training.
    \item \textbf{LLM4CS}~\cite{llm4cs}: A training-free prompting framework that uses an LLM to generate multiple rewrites and hypothetical responses for conversational search.
\end{itemize}
\endgroup

\section{Prompt}
\label{sec:prompt}
\subsection{Rewrite Query and Answer Generation}
Prompt $P_{rewrite}$ rewrites the query and potential answer with given history context. We directly follow the prompts from LLM4CS RAR setting~\cite{llm4cs}.

\subsection{Follow-up Query Generation}
Prompt $P_{follow}$ generates the follow-up query and answer with given history context.
\phantomsection
\begin{tcolorbox}[
  colback=white!95!gray,
  colframe=gray!50!black,
  rounded corners,
  title={\small Follow-up Query and Corresponding Answer}
]
\small
Generate a follow-up query and its answer based on the previous query and answer. \\
The follow-up query should be highly relevant to the topic and context of the conversation.\\ (Closely related so that both questions are likely to be answered by a passage.) \\
If no follow-up is needed, return an empty string. Please format your output exactly as follows: \\
{{"followup\_query": "Your highly related query here", "next\_answer": "Your corresponding answer here"}}\\\\
Context: {History Context}\\
Current Query: {Current Query}\\
        
\end{tcolorbox}

\subsection{LLM Best Doc Selection}
This prompt selects the best 1 document given top-k documents, that best answers the given question and history context.
\phantomsection
\begin{tcolorbox}[
    breakable,
    enhanced,
  colback=white!95!gray,
  colframe=gray!50!black,
  rounded corners,
  title={\small Best Document Selection}
]
\small

You are an expert in question answering.\\
Your task is to select the most relevant passage from the given passages based on the query.\\
I will provide you with {k} passages, each with a passage number.\\[4pt]

Select the passage based on their relevance to the search query: \{question\}.\\
Dialogue History: \{formatted\_context\}\\
Search Query: \{oq\}\\
Passages:\\
\{passage\_text\}\\[6pt]

Please follow the steps below:\\
Step 1: List up the information required to answer the search query and rewrite the query into a standalone and de-contextualized query.\\
Step 2: Analyze whether each passage has the information required to answer the search query.\\
Step 3: Choose the passage that mostly covers clear and diverse information to answer the query.\\[6pt]

The format of the final output should be as follows:\\[4pt]

\{\\
"query analysis": "<step 1 output, your analysis of the query and the rewritten query>",\\
"passage analysis": "<step 2 output, analysis>",\\
"reason for selection": "<step 3 output, reason for selecting the passage>",\\
"selected\_passage\_number": "<the number of the selected passage, chosen from 1 to \{TOP\_K\}>"\\
\}

\end{tcolorbox}

% \phantomsection
% \begin{tcolorbox}[
%   colback=white!95!gray,
%   colframe=gray!50!black,
%   rounded corners,
%   title={\small Best Document Selection}
% ]
% \small
%  You are an expert in question answering. \\
%  Your task is to select the most relevant passage from the given passages based on the query.\\
% I will provide you with 5 passages, each with a passage number. \\
% Select the passage based on their relevance to the search query: {{question}}.\\
% Dialogue History: {formatted\_context}\\
% Search Query: {oq}\\
% Passages:\\
% \{passage\_text\}\\
% Please follow the steps below:\\
% Step 1: List up the information required to answer the search query and rewrite the query into a standalone and de-contextualized query.\\
% Step 2: Analyze whether each passage has the information required to answer the search query.\\
% Step 3: Choose the passage that mostly covers clear and diverse information to answer the query.
% The format of the final output should be as follows:\\
% \{\{\\
% "query analysis": "<step 1 output, your analysis of the query and the rewritten query>",\\
% "passage analysis": "<step 2 output, analysis>",\\
% "reason for selection": "<step 3 output,reason for selecting the passage>",\\
% "selected\_passage\_number": "<the number of the selected passage, chosen from 1 to {TOP_K}>"\\
% \}\}\\
% \end{tcolorbox}

\subsection{LLM Doc Full Reranking}
In Section~\ref{app:llm_reranking}, we compare the performance of LLM-based document selection and full reranking over the top-10 documents. The prompt below corresponds to the latter setting, in which all top-k documents are ranked according to their relevance to the given question and dialogue history.
\phantomsection
\begin{tcolorbox}[
    breakable,
    enhanced,
  colback=white!95!gray,
  colframe=gray!50!black,
  rounded corners,
  title={\small LLM Document Ranking}
]
\small

You are an expert in information retrieval and passage ranking.\\
Your task is to rank the given passages by their relevance to the query.\\
I will provide you with \{TOP\_K\} passages, each indicated by a passage number from 1 to \{TOP\_K\}.\\[4pt]

Rank all passages based on their relevance to the search query.\\
Dialogue History: \{context\}\\
Search Query: \{query\}\\
Passages:\\
\{passage\_text\}\\[6pt]

Please follow the steps below:\\
Step 1: Analyze the search query in the context of the dialogue history and rewrite it into a standalone query.\\
Step 2: Evaluate each passage's relevance to the search query.\\
Step 3: Rank all \{TOP\_K\} passages from most relevant to least relevant.\\[6pt]

The format of the final output should be as follows:\\[4pt]
\{\\
"query\_analysis": "<step 1 output, your analysis and rewritten query>",\\
"passage\_analysis": "<step 2 output, brief relevance analysis of each passage>",\\
"ranking": [<step 3 output, comma-separated passage numbers from most to least relevant, e.g., 3, 1, 5, 2, 4, 7, 6, 8, 10, 9>]\\
\}

\end{tcolorbox}

\section{Usage of GenAI}
AI-assisted coding tools were used to support data analysis and visualization, including drafting and debugging scripts for figure/table generation. All code and results were subsequently reviewed and validated by the authors.

\end{document}